\documentclass[acmtog,authorversion]{acmart}
\acmSubmissionID{502}

\usepackage{booktabs} 

\usepackage{epsfig}
\usepackage{graphicx}
\usepackage{amsmath}
\usepackage{mathrsfs} 
\usepackage{mathtools} 
\usepackage{esdiff} 
\usepackage{amsthm} 
\usepackage{nicefrac}
\usepackage{xfrac}
\usepackage{xcolor} 

\renewcommand{\vec}[1]{\mathbf{#1}}

\DeclareMathOperator{\Var}{Var}

\DeclareMathOperator{\rmse}{RMSE}

\newcommand{\ie}{\textit{i}.\textit{e}., }
\newcommand{\eg}{\textit{e}.\textit{g}., }

\usepackage[ruled]{algorithm2e} 

\SetAlFnt{\small}
\SetAlCapFnt{\small}
\SetAlCapNameFnt{\small}
\SetAlCapHSkip{0pt}

\acmJournal{TOG}
\acmYear{2020}\acmVolume{39}\acmNumber{4}\acmArticle{79}\acmMonth{7} \acmDOI{10.1145/3386569.3392470}

\begin{document}
\title{Quanta Burst Photography}

\author{Sizhuo Ma}
\email{sizhuoma@cs.wisc.edu}
\author{Shantanu Gupta}
\email{sgupta226@wisc.edu}
\affiliation{%
 \institution{University of Wisconsin-Madison}
 \country{USA}}

\author{Arin C. Ulku}
\email{arin.ulku@epfl.ch}
\author{Claudio Bruschini}
\email{claudio.bruschini@epfl.ch}
\author{Edoardo Charbon}
\email{edoardo.charbon@epfl.ch}
\affiliation{%
 \institution{EPFL}
 \country{Switzerland}}

\author{Mohit Gupta}
\email{mohitg@cs.wisc.edu}
\affiliation{%
 \institution{University of Wisconsin-Madison}
 \country{USA}}

\renewcommand\shortauthors{Ma, S. et al}


\begin{abstract}
Single-photon avalanche diodes (SPADs) are an emerging sensor technology capable of detecting individual incident photons, and capturing their time-of-arrival with high timing precision. While these sensors were limited to single-pixel or low-resolution devices in the past, recently, large (up to 1 MPixel) SPAD arrays have been developed. These single-photon cameras (SPCs) are capable of capturing high-speed sequences of binary single-photon images with no read noise. We present quanta burst photography, a computational photography technique that leverages SPCs as \emph{passive imaging devices} for photography in challenging conditions, including ultra low-light and fast motion. Inspired by recent success of conventional burst photography, we design algorithms that align and merge binary sequences captured by SPCs into intensity images with minimal motion blur and artifacts, high signal-to-noise ratio (SNR), and high dynamic range. We theoretically analyze the SNR and dynamic range of quanta burst photography, and identify the imaging regimes where it provides significant benefits. We demonstrate, via a recently developed SPAD array, that the proposed method is able to generate high-quality images for scenes with challenging lighting, complex geometries, high dynamic range and moving objects. With the ongoing development of SPAD arrays, we envision quanta burst photography finding applications in both consumer and scientific photography.
\end{abstract}

%
%


\begin{teaserfigure}
  \includegraphics[width=0.97\linewidth]{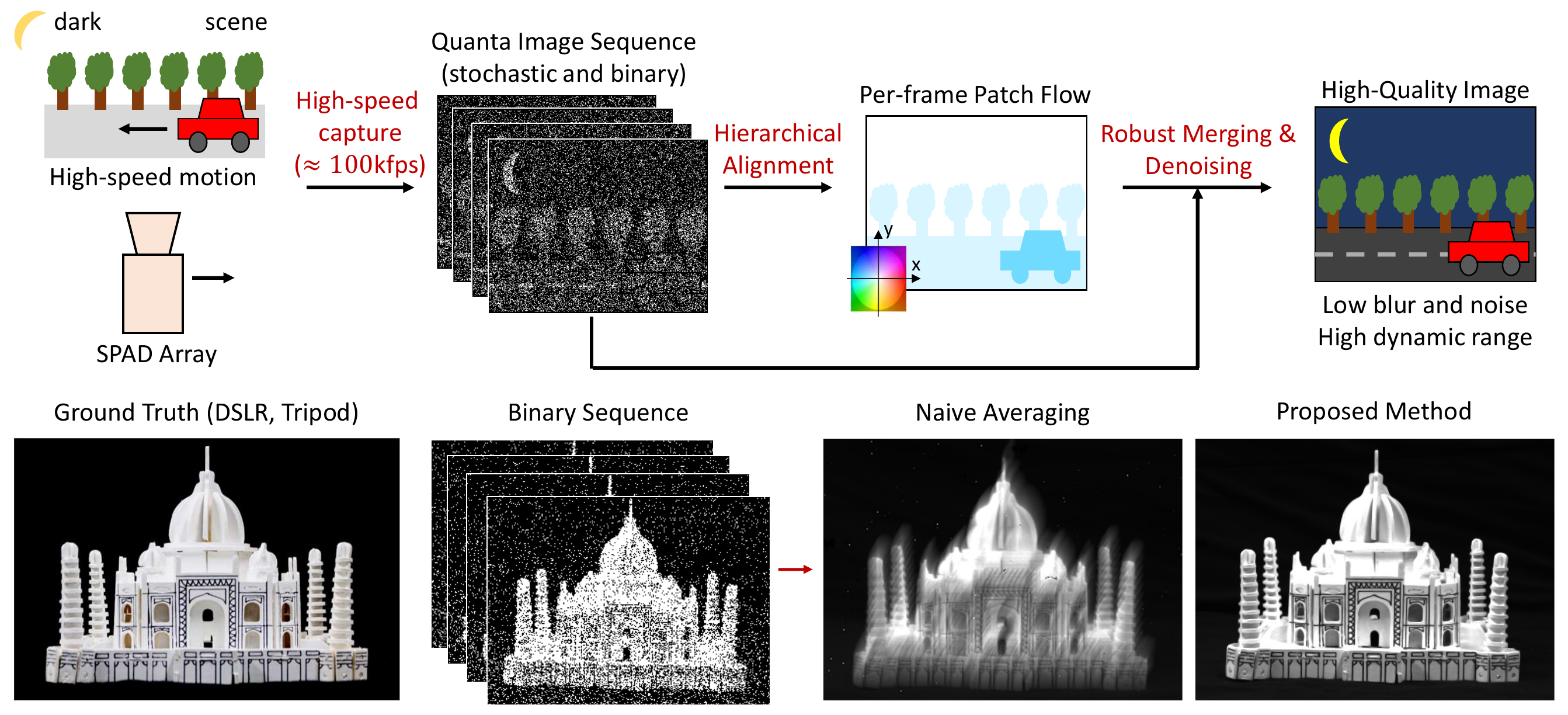}
  \vspace{-0.0in}
  \caption{\textbf{Quanta burst photography.} (Top) Single-photon image sensors capture stochastic, binary image sequences at high speeds ($\sim 100$ kfps). Such high-speed image sequences can be aligned to compensate for scene/camera motion using a spatial-temporal hierarchical alignment algorithm. By merging the aligned sequence robustly, a high-quality image can be reconstructed, with minimal motion blur and noise, and high dynamic range, even in challenging photography conditions. (Bottom, from left to right) An example low-light scene captured by a DSLR camera on a tripod to avoid camera shake; binary image sequence captured by a handheld single-photon camera; image reconstructed by naive averaging of the binary sequence (shown to illustrate the amount of motion during capture); super-resolved image reconstructed using the proposed techniques has low blur and noise. {\bf Zoom in for details.} }
  \label{fig:teaser}
  \vspace{-0.0in}
\end{teaserfigure}

\maketitle

\section{The Single-Photon Revolution}
A conventional camera typically captures hundreds to thousands of photons per pixel to create an image. An emerging class of sensors, called single-photon avalanche diodes (SPADs)~\cite{rochas_spad_phd_2003,niclass_design_2005}, can record \emph{individual} photons, and precisely measure their time-of-arrival. Due to their sensitivity and picosecond time resolution, SPADs are driving an imaging revolution. A new generation of devices is emerging, with novel functionalities such as imaging at trillion fps~\cite{OToole:2017:SPAD}, non-line-of-sight (NLOS) imaging~\cite{buttafava2015non,OToole2018}, and microscopic imaging of nano time-scale bio-phenomena~\cite{bruschini_single-photon_2019}.


\paragraph{Passive single-photon imaging:} So far, most SPAD-based imaging systems are active, where the SPAD is used in precise temporal synchronization with an active light source (e.g., a pulsed laser). This includes applications such as NLOS imaging, LiDAR~\cite{Shin_2016_naturecomm}, and microscopy. Can SPADs be used not just with controlled and precisely synchronized active light sources as is the norm, but more generally under passive, uncontrolled illumination (e.g., sunlight, moonlight)? Such passive SPAD-based imaging systems have the potential to expand the scope of SPADs to a considerably larger set of applications, including machine vision and photography. 

Consider a SPAD sensor (an array of SPAD pixels) imaging a scene illuminated by passive lighting. Since photons arrive at the sensor randomly according to Poisson statistics, photon detection events are also random, and can be visualized as a spatio-temporal \emph{photon-cube}~\cite{fossum_quanta_2011}. A SPAD camera can capture a sequence of thin, temporal slices of the photon-cube, where each slice is a binary (1-bit) image, as shown in in Fig.~\ref{fig:teaser}. Each pixel location records a $1$ if it receives one or more photons during the temporal extent of the slice, and $0$ otherwise. For example, a recent SPAD camera~\cite{ulku_512_2019} can capture $\sim 10^5$ binary frames per second, at 1/4 MPixel resolution.\footnote{Photon-cubes and single-photon binary image sequences were first considered in the context of jots~\cite{fossum_what_2005,fossum_quanta_2011}, another emerging single-photon sensing technology. In this paper, we primarily focus on SPADs due to their high frame rate. However, since both jots and SPADs have similar imaging model and data format, the analysis and techniques presented here are applicable to jots as well.} Due to the random nature of photon arrivals, the binary images are stochastic.



\paragraph{Passive single-photon imaging under motion:} How does motion manifest in a stochastic binary image sequence? If the scene (or camera) moves during acquisition, the photons emitted by a scene point get \emph{mis-aligned} and spread over multiple SPC pixels. In this paper, we propose \emph{quanta burst photography}, a computational photography technique that computationally re-aligns the photons along motion trajectories, for achieving high-quality images in challenging scenarios, including low-light and high-speed motion  (Fig.~\ref{fig:teaser}). We develop algorithms that align the binary slices, thus creating a high-bit-depth, high-dynamic-range, potentially super-resolved (via sub-pixel alignment~\cite{park_super-resolution_2003,wronski_handheld_2019}) image of the scene, while minimizing noise and motion blur. This is similar in spirit to conventional burst photography where a burst of noisy, short-exposure images are aligned and merged into a single high-quality image~\cite{hasinoff_burst_2016,liba_handheld_2019}. Quanta burst photography can be considered a limiting case because each binary image captures at most one photon per pixel, and is thus extremely noisy and quantized (1-bit). On the other hand, due to fast capture, we have a long sequence available ($10^2-10^5$ frames, depending on light level, dynamic range and motion), instead of $5-10$ as in conventional burst photography. 

\paragraph{Why quanta burst photography?} One of the key benefits of SPCs is the low read noise in the raw binary frames~\cite{bruschini_single-photon_2019}, which enables dividing the exposure time finely into a long sequence of frames to handle fast motion. This results in virtually negligible intra-frame motion blur and low noise, even for rapid motion (e.g., sports and wildlife photography).~\footnote{For conventional cameras, there is a fixed read noise penalty for each captured frame. Therefore, dividing the exposure time finely into a large number of frames increases the effective read noise in the merged image.} Furthermore, although at first glance it may appear that SPCs, due to their high sensitivity, are useful only in photon-starved scenarios, surprisingly, they can also image bright scenes where conventional sensors saturate~\cite{ingle_high_2019,antolovic_dynamic_2018}. This is because although each binary image is quantized, a large collection of single-photon measurements, when combined, naturally avoids saturation~\cite{yang_bits_2012}, and thus, achieve extreme dynamic range. There are two catalysts for key quanta burst photography:


\subparagraph{(a) Emergence of large SPCs arrays:} Till recently, SPCs were available as single-pixel or small arrays (e.g., 32x32 pixels), which, while sufficient for several scientific imaging applications, are not suitable for consumer domain photography. Fortunately, due to their compatibility with mainstream CMOS fabrication lines, it is now possible to develop large SPCs arrays, with the world's first 1 MPixel jots~\cite{ma_photon-number-resolving_2017} and SPAD arrays~\cite{Morimoto:2019} reported recently, while maintaining high sensor quality \emph{and room temperature operation}.


\subparagraph{(b) High-performance burst photography:} We are inspired by the recent success of burst photography algorithms \cite{hasinoff_burst_2016,wronski_handheld_2019,liba_handheld_2019}, which, for the first time, are starting to produce reliably artifact-free images in almost all circumstances, including challenging scenes with occlusions and non-rigid motion. These motion estimation and merging methods are robust enough to be shipped to consumer devices, a gold-standard for computational photography techniques. 

We adopt the design principles and best practices from these burst photography approaches, and design algorithms tailored for single-photon binary stochastic images. We demonstrate, via simulations and experiments on a 1/8 megapixel SPAD array (SwissSPAD2~\cite{ulku_512_2019}) that quanta burst photography is able to generate high SNR, blur-free and super-resolved images in extreme scenarios (low-light, fast motion, large dynamic range) which would be considered challenging for burst photography on conventional cameras.

\paragraph{Scope and limitations:} Are single-photon cameras and quanta burst photography ready to be deployed on consumer devices? Not yet. So far, we have focused on achieving high image quality. Our current unoptimized implementation, however, is not directly amenable to consumer devices, which have strong constraints on speed, power and memory. The current sensor prototype does not have a color filter array (e.g., a Bayer pattern), and thus, the resulting images are gray-scale. The resolution, although highest to-date among SPAD cameras, is still relatively low (1/8 MPixel) for consumer applications. Fortunately, the capabilities of single-photon sensors continue to improve, with higher resolution~\cite{Morimoto:2019} and color sensors~\cite{Elgendy:2019} on the horizon. 


The proposed approach is not meant to directly compete with conventional CMOS sensors and burst photography pipelines, which have been optimized over several years, and can produce compelling photographic experiences. Instead, our goal is to explore and analyze a nascent but promising imaging modality, which, if successful, could lead to capabilities (high quality photography in ultra low-light and fast motion) that were hitherto considered impossible. This paper should be seen just as a first step toward that goal. 

\section{Related Work}
\paragraph{Image denoising.} Single image denoising algorithms denoise images by imposing various image prior such as spatial smoothness \cite{beck_fast_2009}, sparsity~\cite{elad_image_2006} and self-similarity~\cite{buades_non-local_2005}. Such priors are also used in transform domains such as frequency domain~\cite{gonzalez_digital_2006}, wavelets~\cite{malfait_wavelet-based_1997}, and 3D transform (BM3D)~\cite{dabov_image_2007}. Recent data-driven denoising algorithms attempt to capture the noise statistics using a neural network instead of an explicit prior~\cite{zhang_ffdnet_2018}. Single image denoising approaches tend to fail when the image has a very low SNR due to low light, and/or limited exposure time because of fast scene/camera motion. In these cases it is essential to combine information from multiple images to generate a high-quality image.


\paragraph{Burst denoising.} Burst denoising methods take a sequence of underexposed images and merge them into a single image. The SNR is improved since more photons are collected. The key technical challenge for burst denoising is to accurately align and merge frames as the camera moves. This can be addressed either by a two-step align-and-merge approach~\cite{wronski_handheld_2019,liba_handheld_2019,hasinoff_burst_2016,liu_fast_2014,heide_flexisp_2014}, or joint optimization~\cite{heide_proximal_2016}. Recently, deep learning based methods have also been proposed~\cite{godard_deep_2018,mildenhall_burst_2018}. A related problem is video denoising, where both input and output are sequence of images~\cite{chen_seeing_2019,maggioni_video_2012,dabov_video_2007}. Our goal is to create a single, high-quality image from a burst of binary single-photon frames. 




\paragraph{Quanta (single-photon) sensors.} Currently, there are two main enabling technologies for large single-photon camera arrays: SPADs and jots. SPADs achieve single photon sensitivity by amplifying the weak signal from each incident photon via avalanche multiplication, which enables zero read noise and extremely high frame rate ($~\sim 100$kfps). Jots, on the other hand, amplify the single-photon signal by using an active pixel with high conversion gain (low capacitance)~\cite{fossum_what_2005}. By avoiding avalanche, jots achieve smaller pixel pitch, higher quantum efficiency and lower dark current, but have lower temporal resolution~\cite{ma_photon-number-resolving_2017}. Although the techniques in this paper are applicable to both SPADs and jots, we primarily focus on SPADs because of their capability to resolve fast motion due to high temporal resolution. We show simulation-based comparisons between the two types of sensors, and a discussion on their relative merits, in Sec.~\ref{sec:results}.

\paragraph{Wide-dynamic-range sensors.} There are several wide-dynamic-range image sensors based on different technologies, such as logarithmic response~\cite{kavadias_logarithmic_2000} and light-to-frequency conversion~\cite{wang_high_2006}. Such sensors have an extended dynamic range compared to conventional CMOS sensors, but the blur-noise trade-off still exists, which makes them less effective for low-light, fast-motion scenarios. In addition, especially for logarithmic sensors, photo response non-uniformity (PRNU) is a  limitation in conventional implementations~\cite{yang_image_2009}; this effect compounds the above issues, significantly limiting image quality.


\paragraph{Image reconstruction from single-photon sensor data.} There is prior work on reconstructing intensity images from single-photon binary frames using denoising techniques such as total variation and BM3D \cite{chan_images_2016,gnanasambandam_megapixel_2019}, or by an end-to-end neural network~\cite{choi_image_2018,chandramouli_bit_2019}. In the presence of motion, Fossum \shortcite{fossum_modeling_2013} suggested shifting the binary images to compensate for motion and achieve blur-free image reconstruction. This idea has been implemented recently~\cite{iwabuchi_iterative_2019,gyongy_single-photon_2018,gyongy_object_2017}, albeit for simplistic motion models (\eg planar objects with in-plane motion and no occlusions). Our approach is based on a much less restrictive assumption (motion can be approximated by patch-wise 2D translation and remains constant within temporal blocks), and can reliably produce high-quality images for a broad range of complex, real-world scenes.


\section{Passive Single-Photon Imaging Model} \label{sec:math}
Consider a SPC pixel array observing a scene. The number $Z (x, y)$ of photons arriving at pixel $(x, y)$ during an exposure time of $\tau$ seconds is modeled as a Poisson random variable~\cite{yang_bits_2012}:
\begin{equation} \label{eq:poisson}
    P\{Z=k\} = \frac{(\phi\tau\eta)^{k}e^{-\phi\tau\eta}}{k!}\,,
\end{equation}
where $\phi (x,y)$ is the photon flux (photons/seconds) incident at $(x, y)$. $\eta$ is the quantum efficiency. Each pixel detects at most one photon during an exposure time, returning a binary value $B(x,y)$ such that $B(x,y) = 1$ if $Z(x,y) \geq 1$; $B(x,y) = 0$ otherwise.
Due to the randomness in photon arrival, $B(x,y)$ is a random variable with Bernoulli distribution:
\begin{equation}\label{eq:bernoulli}
    \begin{aligned}
        P\{B=0\} &= e^{-(\phi\tau\eta+r_q\tau)} \,,  \\
        P\{B=1\} &= 1-e^{-(\phi\tau\eta+r_q\tau)} \,,
    \end{aligned}
\end{equation}
where $r_q$ is the dark count rate (DCR), which is the rate of spurious counts unrelated to photons. 

To estimate the number of incident photons $\phi$ (proportional to the linear intensity image of the scene), suppose the camera captures a sequence of binary frames. Assuming no motion between binary frames, or that the binary frames are aligned perfectly to compensate for motion, we define $S(x,y)$ as the sum of all binary frames:
\begin{equation} \label{eq:sum_image}
    S(x,y) = \sum_{t=1}^{n_q} B_t(x,y)\,,
\end{equation}
where $B_t(x,y)$ is the binary frame at time $t$, and $n_q$ is the number of frames. $S (x,y)$ is the total number of photons detected at $(x,y)$ over the entire binary image sequence. Since each binary frame is independent, the expected value of the sum image is the product of the number of frames $n_q$, and the expected value of the Bernoulli variable $B$:
\begin{equation} \label{eq:sum_image_expectation}
    E [S(x,y)] = n_q \, E [B(x,y)] = n_q \left( 1-e^{-(\phi\tau\eta + r_q\tau)} \right).
\end{equation}

The maximum likelihood estimate (MLE) of the intensity image $\phi$ is given as~\cite{antolovic_nonuniformity_2016}:
\begin{equation} \label{eq:mle}
    \hat{\phi}(x,y) = -\ln (1-S(x,y)/n_q)/\tau\eta-r_q(x,y)/\eta\,.
\end{equation}

\begin{figure}
  \includegraphics[width=\linewidth]{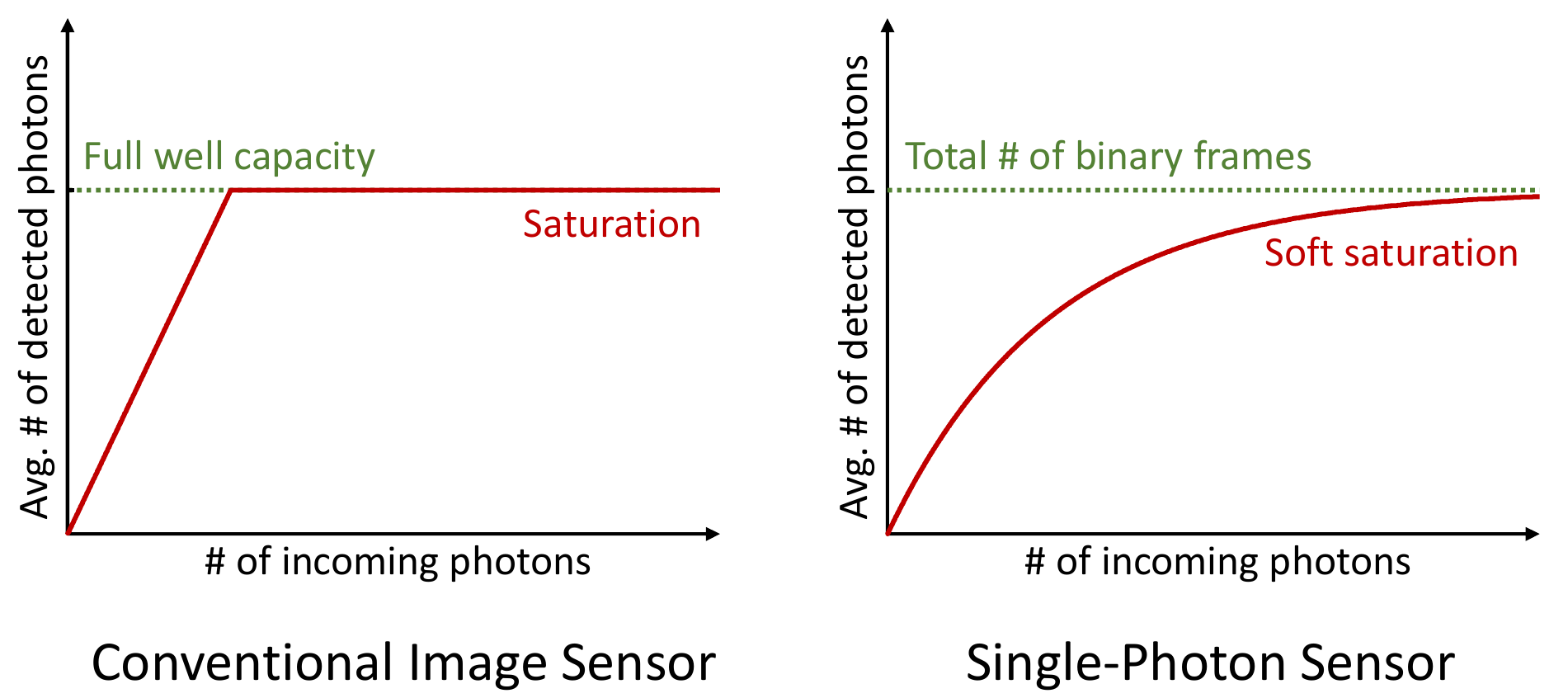}
\vspace{-0.2in} \caption{\textbf{Response curves for conventional sensors and SPADs.} The response curve for a sensor is defined as the plot of the average number of photons detected as a function of number of photons incident on the sensor. \textbf{(Left)} The response curve for conventional sensors is linear, until saturation when the full well capacity is reached. \textbf{(Right)} For SPADs, the response curve is non-linear, and asymptotically approaches a limit, which is the total number of binary frames captured in the given time duration. SPADs suffer from only soft saturation since the number of detected photons keeps increasing, albeit progressively slowly, for increasing incident flux.}
\vspace{-0.2in} \label{fig:response_curves}
\end{figure}

\begin{figure*}
  \includegraphics[width=0.96\linewidth]{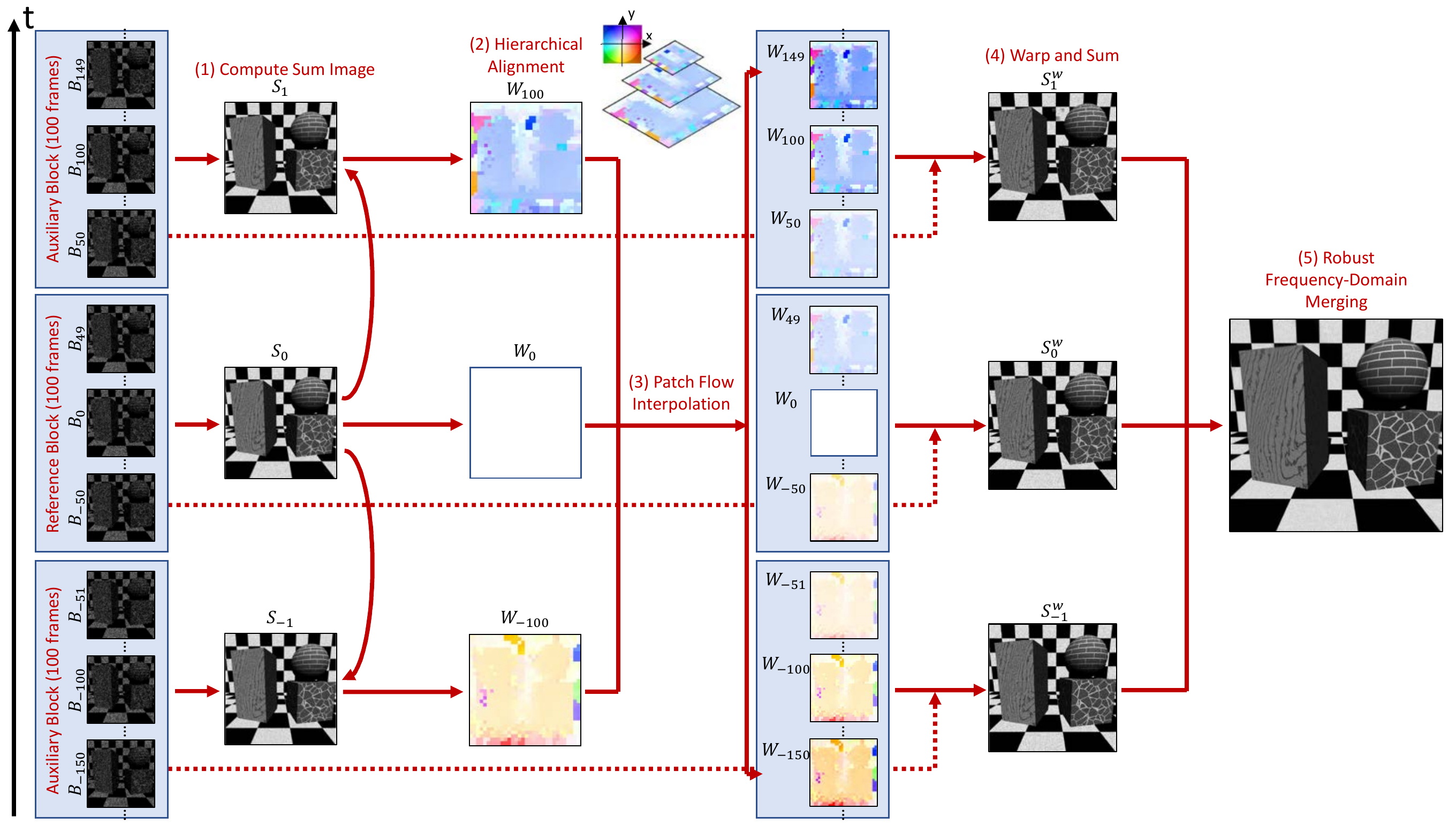}
 \vspace{-0.0in} \caption{\textbf{Algorithm overview.} In this example, the binary sequence is divided into 100-frame temporal blocks. The central block is chosen as the reference block. \textbf{(1)} For each block, the binary frames are added to form the block-sum image. \textbf{(2)} Every other block is aligned to the reference block, resulting in a coarse patch flow between the center frames of the blocks. \textbf{(3)} Coarse patch flow is temporally interpolated to estimate the fine-scale patch flow between individual binary frames. \textbf{(4)} Binary frames are warped using the fine-scale patch flow and added together to form warped block-sum images. \textbf{(5)} Warped block-sum images are merged together using a robust frequency-domain approach.} \vspace{-0.1in}\label{fig:algorithm}
\end{figure*}

\paragraph{Dynamic range:} Eq.~\ref{eq:sum_image_expectation} describes the relationship between $S$, the total number of photons detected by the camera, and $\phi$, the number of photons incident on the camera (the quantity we wish to estimate). This non-linear relationship~\cite{sbaiz_gigavision_2009}, as plotted in Fig.~\ref{fig:response_curves}, is similar to the D-log H curve for photographic films proposed by Hurter and Diffield in 1890, because single-photon cameras emulate the silver halide emulsion film process~\cite{fossum_what_2005}. The key observation is that this \emph{response curve} asymptotically approaches its maximum value ($n_q$), while never reaching it. This \emph{soft saturation}~\cite{ingle_high_2019} suggests that the value of $S$ keeps increasing (albeit progressively slowly) as the number of incident photons increases, which means the incident flux can be recovered even for bright scenes. In contrast, the response curve for conventional sensors is a straight line before hitting the full well capacity, and then flattens due to saturation. Therefore, a passive single-photon camera, while capable of imaging low-light scenes, somewhat counter-intuitively, can also image bright scenes where conventional sensors saturate, providing an extremely wide dynamic range. 


\paragraph{Read noise:} Conventional sensors convert discrete incident photons to analog current, which is again converted to a discrete number by an analog-to-digital converter (ADC). This discrete $\rightarrow$ analog $\rightarrow$ discrete pipeline results in read noise, which is the dominant source of noise in low-light. This places a limit on exposure time used in conventional burst photography. Given a fixed total capture time, increasing the number of frames may reduce motion artifacts, but since each additional frames incurs a read noise penalty, the SNR of the merged image is lowered. Jots have a deep sub-electron read noise (currently $\sim$ 0.2e$^-$~\cite{ma_photon-number-resolving_2017}), which although considerably lower than conventional CMOS sensors, can still limit the image quality in ultra low-light conditions~\cite{fossum_quanta_2016}.

In contrast, SPADs \emph{directly} measure the photon counts, skipping the intermediate analog conversion, thereby avoiding read noise. This allows a SPAD camera to finely divide the exposure time into a large number $n_q$ of binary frames for motion compensation, thereby simultaneously achieving low motion-blur and high SNR.

\section{Single-Photon Imaging Under Motion} \label{sec:overview}
If the scene or camera moves during capture, then simply summing the binary sequence (Eq.~\ref{eq:sum_image}) leads to merging of photons from different scene points, resulting in motion blur. Therefore, to avoid motion blur, the binary frames must be aligned to compensate for inter-frame motion before merging them. 


Aligning the binary frames directly is challenging because the traditional brightness constancy assumption does not hold for the observed random binary signal due to extremely low SNR. Although it may be possible to estimate the inter-frame motion when the motion is a global, low-dimensional transform such as global 2D translation or global homography, for general, unstructured scenes with unknown geometry, the transform must be formulated as a pixelwise 2D motion field (or optical flow). In this case, the total number of unknown parameters to estimate is $2MN$ for image resolution $M\times N$. Such a complex, high-dimensional motion model cannot be solved precisely from the random binary input data.

Fortunately, SPADs are able to capture binary frames at high frame rates (97.7kfps for SwissSPAD2~\cite{ulku_512_2019}). At such high frame rates, the velocity at each pixel can be treated as a constant within a local temporal window. We use this observation as an additional constraint to solve the otherwise challenging optical flow problem on stochastic binary frames. One way to incorporate such a constraint is to compute a temporally coherent optical flow~\cite{black_recursive_1994,weickert_variational_2001,volz_modeling_2011}. In practice, we choose a simple, less computationally intensive approach: We divide the entire image sequence into non-overlapping temporal blocks, compute the sum image for each block (called \emph{block-sum images}) and align the block-sum images. The block-sum images have a higher SNR than individual binary frames, which makes it possible to use traditional optical flow methods to align them. \smallskip



\paragraph{Block-level vs. frame-level alignment.} Fig.~\ref{fig:algorithm} shows an overview of the method. We call the block in the center of the sequence the \emph{reference block}. All the other blocks, called \emph{auxiliary blocks}, are aligned to the reference block. After aligning the block-sum images, we do not use the coarse-temporal-scale motion field between temporal blocks to merge them directly. Instead, we linearly interpolate the motion field in time to obtain motion between successive binary frames. This fine-scale motion field is used to warp each binary frame and align to a central reference frame in the reference block, before merging. This hierarchical approach removes the motion blur within each temporal block, resulting in sharp images even for fast moving scenes. After warping, a frequency-space merging algorithm is used to merge the temporal blocks, which provides robustness to small alignment error. In the next two sections, we provide details of the align and merge algorithms.


\section{Aligning Temporal Blocks} \label{sec:align}
Given a reference and an auxiliary block, we compute the 2D correspondence map between them based on their appearance. Instead of using a pixel-wise optical flow algorithm, we use a patch-based alignment approach~\footnote{In this paper, we refer to temporal sum of frames as ``blocks'' and spatial windows of pixels as ``patches''.} since it is more resilient to noise than pixel-wise optical flow~\cite{bruhn_lucas/kanade_2005,zimmer_optic_2011}. Furthermore, even for merging (Sec.~\ref{sec:merge}), patch-based approaches achieve more robust results than pixel-based merging~\cite{liu_fast_2014} in low SNR images. For patch-based merging, it is sufficient to compute a motion field at the patch level, thereby saving computational time. 


\paragraph{Hierarchical patch alignment.} We use a hierarchical patch alignment approach similar to \cite{hasinoff_burst_2016} on an image pyramid built from the block-sum images. The number of pyramid levels can be adjusted according to the spatial resolution of the binary images. We use a 3-level pyramid for the 512x256 images used in our experiments. The matching is done by minimizing L1 matching error in a spatial neighborhood. For a patch with indices $(i,j)$, which expands the pixel indices $[iM,(i+1)M-1]\times[jM,(j+1)M-1]$, we find the smallest motion vector $(u,v)$ that minimizes:
\begin{equation} \label{eq:matching_error}
    E_d(u,v;i,j) = \sum_{x=iM}^{(i+1)M-1}\sum_{y=jM}^{(j+1)M-1}|S_{aux}(x+u,y+v)-S_{ref}(x,y)| \,.
\end{equation}
The size of the patch is $M\times M$. $S_{aux}$ is the auxiliary block-sum image and $S_{ref}$ is the reference block-sum image.

\paragraph{Spatial regularization at finest level. } We perform a global regularization at the finest level of the pyramid to further refine the patch alignment results (especially for blocks with extremely small number of photons) and to provide sub-pixel alignment for super-resolution. This is performed by minimizing the following energy:
\begin{equation}
    \min_{\vec{u},\vec{v}}E(\vec{u},\vec{v}) = \int_{\Omega_{ij}}E_d(\vec{u},\vec{v};i,j)+\lambda(\|\nabla\vec{u}\|_1+\|\nabla\vec{v}\|_1)\,didj \,,
\end{equation}
where $\Omega_{ij}=[0,W/M]\times[0,H/M]$ is the spatial domain for the patch indices $i,j$. $\vec{u},\vec{v}$ are the motion fields defined on $\Omega_{ij}$, and $H\times W$ is the spatial resolution of the input images. $E_d$ is the matching error defined in Eq.~\ref{eq:matching_error}. In practice, we minimize the Charbonnier loss $\rho(x)=\sqrt{x^2+\epsilon^2}$ as an differentiable alternative for the L1 loss.



\begin{figure}
  \includegraphics[width=\linewidth]{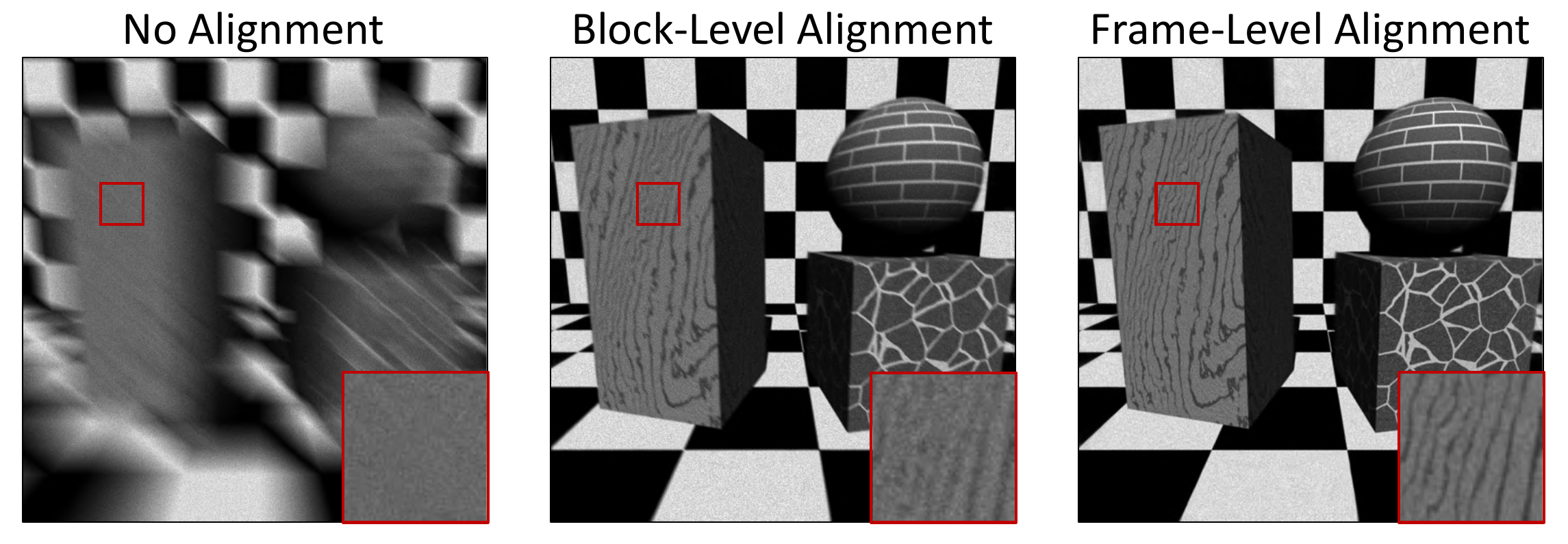}
 \vspace{-0.3in}  \caption{\textbf{Effect of frame-level alignment.} \textbf{(Left)} Simple sum of binary frames captured by a moving camera shows the significant motion blur. \textbf{(Center)} Alignment at the block level (each consisting of 100 binary frames) does not remove the motion blur completely. \textbf{(Right)} Blur is reduced by interpolating the block-level alignment to achieve frame-level alignment.}
  \label{fig:remove_motion_blur}
\vspace{-0.1in}\end{figure}

\paragraph{Interpolating the motion field.} The computed inter-block motion is treated as motion between the center frames of each block. A linear interpolation is then performed to compute the motion between individual frames. Higher-order interpolation (\eg cubic or spline) may improve the temporal smoothness, but will increase the dependency on other blocks. In practice, linear interpolation achieves good results for SPADs with high temporal resolution. Fig.~\ref{fig:remove_motion_blur} shows an example demonstrating the benefits of frame-level interpolation. While alignment at the block level does not remove the motion blur completely, blur is considerably reduced by frame-level interpolation. An evaluation of frame-level interpolation on real data is provided in the supplementary report.

\section{Merging Binary Sequence} \label{sec:merge}
After estimating inter-frame motion, one way to merge the binary image sequence is to warp the binary images, compute the sum image of all warped images, and finally, compute the MLE of the sum (Eq.~\ref{eq:mle}). However, the estimated motion field may have errors due to occlusions, motion discontinuities, and non-rigid scene deformations. In this case, simply summing the warped binary images will create strong blurring or ghosting artifacts.

\begin{figure}
  \includegraphics[width=\linewidth]{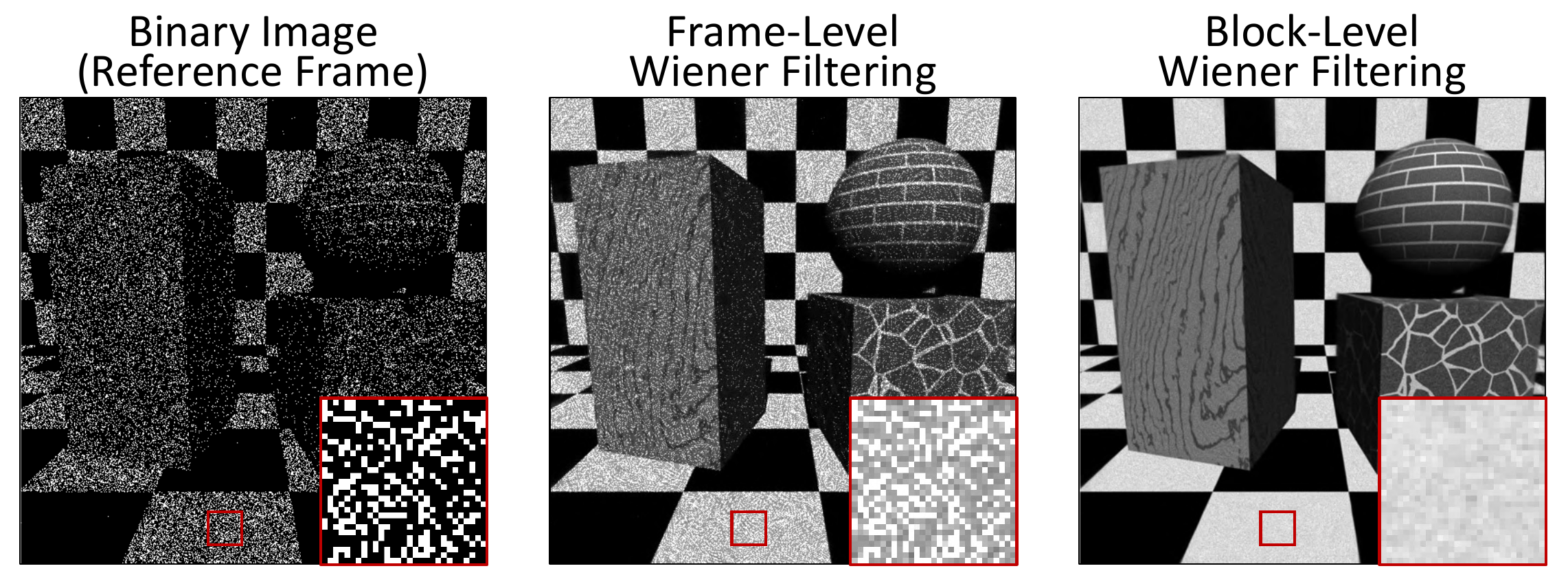}
 \vspace{-0.25in} \caption{\textbf{Block-level Wiener filtering.} \textbf{(Left)} The binary reference frame is extremely noisy due to the stochastic nature of photon arrival. \textbf{(Center)} Wiener filtering is applied such that each auxiliary frame is weighted by measuring its difference with the reference frame. Since the difference is large even for mid and low spatial frequencies, the noise in the reference frame is preserved in the merged image. \textbf{(Right)} Wiener filtering is applied to warped block-sum images, resulting in merged images with higher SNR.}
  \label{fig:wiener_block}
\vspace{-0.1in}\end{figure}

\paragraph{Can robust merging be used for binary frames?} Robust merging methods such as Wiener frequency-domain filtering have long been used in video denoising and burst denoising~\cite{hasinoff_burst_2016} to account for potentially incorrect estimated motion. The key idea is that if a patch in a warped frame is significantly different from that in the reference frame, then the alignment is likely erroneous. The final merged patch is computed by taking a weighted average of all matched patches, where the patches with large difference with the reference path (likely erroneous) are given a lower weight. This approach, while successful for conventional cameras, cannot be directly applied to merge the single-photon binary frames. This is because even if two binary frames are \emph{perfectly aligned}, the difference between the frames could still be high due to the dominating shot noise. As a result, every auxiliary frame will have a low weight, and will make a low contribution to the final merged image, resulting in low SNR, as shown in Fig.~\ref{fig:wiener_block}. 

In order to address this limitation, we adopt a two-step approach. First, we warp the frames within each block to the block's reference frame by using the estimated fine-scale inter-frame motion. The frames are simply added to form a warped block-sum image without any robust merge, since the amount of motion within each block is small, reducing the likelihood of alignment errors. This warping makes it possible to remove the motion blur within each block, as shown in Fig.~\ref{fig:remove_motion_blur}. The warped block-sum images have sufficient SNR to be amenable to a traditional frequency-domain robust-merging approach~\cite{hasinoff_burst_2016}. Therefore, Wiener filtering is applied to the warped block-sum images in the second step, so that they can be merged stably to reduce the noise level. Fig.~\ref{fig:wiener_block} shows the result of applying block-level Wiener filtering, resulting in considerably higher SNR than naive frame-level merging. \smallskip

\begin{figure}
  \includegraphics[width=\linewidth]{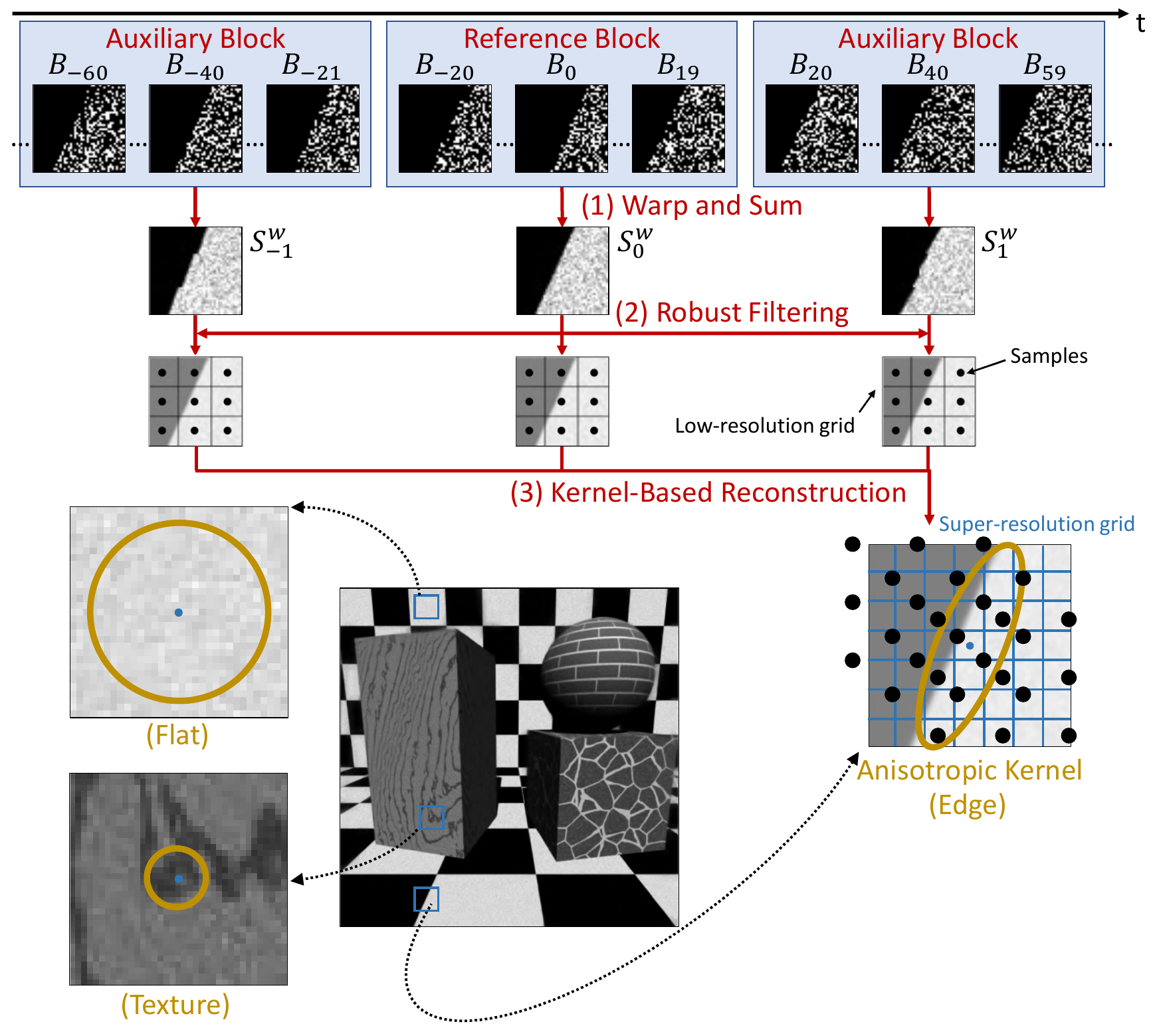}
  \vspace{-0.2in}\caption{\textbf{Super-resolution merging.} \textbf{(1)} Binary frames within a block are warped and summed using the fine-scale inter-frame patch flow. \textbf{(2)} The resulting warped block-sum image is filtered according to a guide image (the warped block-sum image for the reference block). This step prepares matches for the reconstruction step by mitigating the noise and alignment errors. \textbf{(3)} The weighted patches are placed on a supersampled output grid, where pixels in the individual patches are treated as samples. For each pixel on the output grid, an anisotropic Gaussian kernel is used to combine the samples in a local neighborhood. The shape and size of the anisotropic kernel is determined by analyzing the structure tensor of the guide image.}
  \vspace{-0.1in}\label{fig:sr}
\end{figure}
\paragraph{Merging with super-resolution:} The high-speed single-photon data leads to small inter-frame motion ($\sim 0.01$ pixels), which can be leveraged to generate a merged image that has a higher resolution than the input frames~\cite{park_super-resolution_2003,wronski_handheld_2019}. We develop a simple super-resolution algorithm based on kernel regression by adapting the original merging method described above. As above, after the inter-frame motion field is computed, frames within the same block are warped and added up to form the warped block-sum images. However, instead of computing the weighted average of patches, the weighted patches are treated as a bag of sample points, as shown in Fig.~\ref{fig:sr}. Each patch is warped to sub-pixel locations on a higher-resolution output pixel grid. The algorithm then scans through each pixel on the output grid. At each pixel, an anisotropic Gaussian kernel~\cite{takeda_kernel_2007,wronski_handheld_2019} is used to combine the sample points within a spatial neighborhood. Instead of the point-wise robustness term used in recent conventional burst photography~\cite{wronski_handheld_2019}, our super-resolution method uses the frequency-domain robust merging approach (the same approach used in original-resolution merging). This approach is more robust in practice, at the cost of slightly higher computational complexity. Please refer to the supplementary technical report for design details of the kernel regression method.

\paragraph{Post-denoising and tone mapping:} After using the proposed motion-compensating temporal denoising method to generate a final sum image, existing single-photon image reconstruction methods can be applied for further denoising (see the supplementary report for comparisons of different reconstruction methods). We apply Anscombe transform~\cite{anscombe_transformation_1948} to the sum image and apply BM3D~\cite{dabov_image_2007} for spatial denoising~\cite{chan_images_2016}. 

After merging and denoising, we use Eq.~\ref{eq:mle} to invert the non-linear response to get a linear image. Gamma correction and tone-mapping is then applied to generate images suited for viewing.


\begin{figure*} 
  \includegraphics[width=0.96\linewidth]{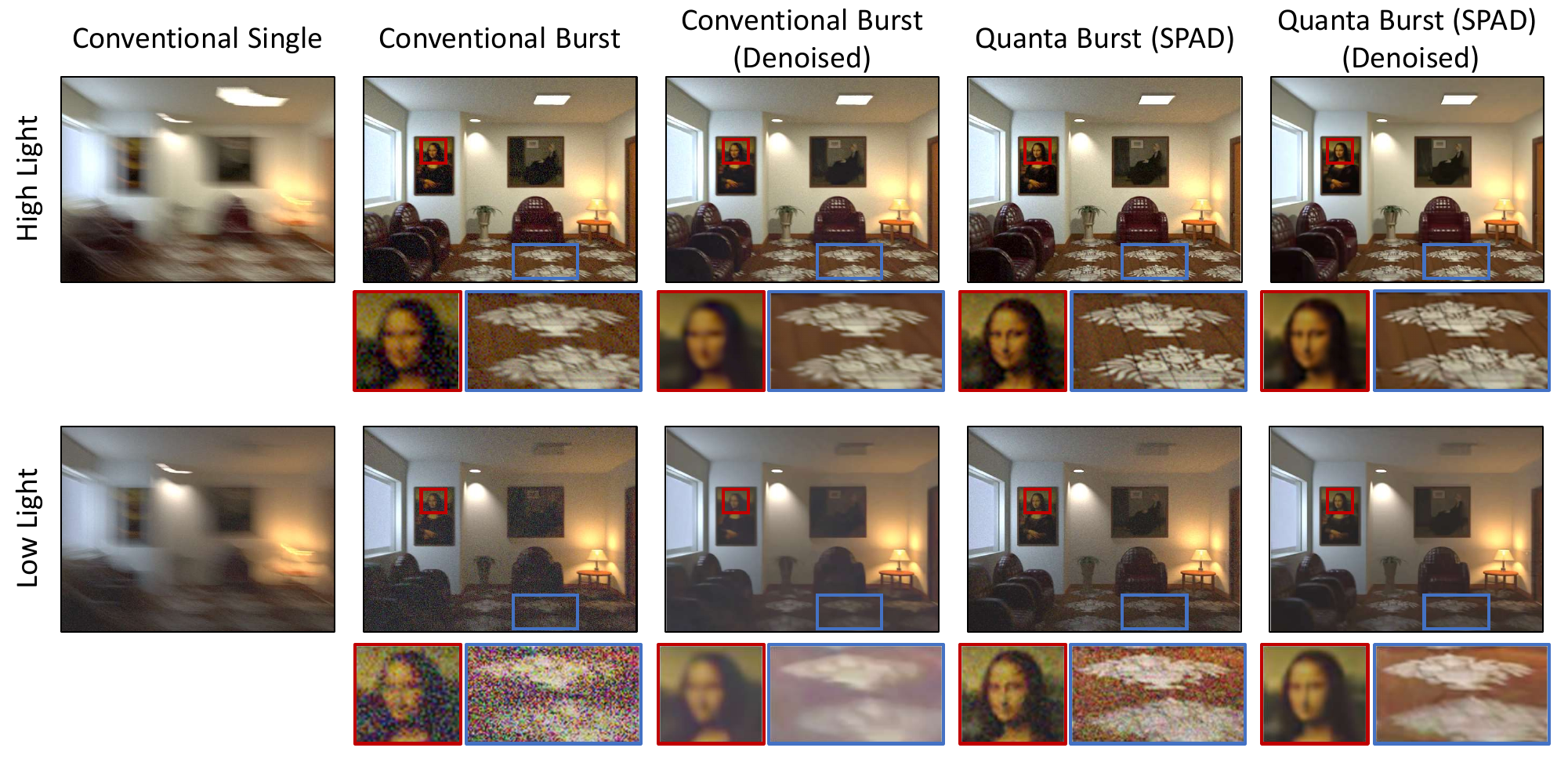}
 \vspace{-0.1in} \caption{\textbf{Simulation results under different lighting conditions.} We simulate a 2000-frame binary sequence of a still indoor scene under three different lighting conditions. The camera motion is the same in all three sequences. \textbf{(Top)} When there is sufficient light, the SNR of conventional and quanta burst photography are comparable, although the latter generates a sharper image with less motion blur. \textbf{(Bottom)} As the light level decreases, quanta burst provides a higher SNR than conventional cameras. }
 \vspace{-0.0in} \label{fig:sim_lighting}
\end{figure*}

\begin{figure*} 
  \includegraphics[width=0.96\linewidth]{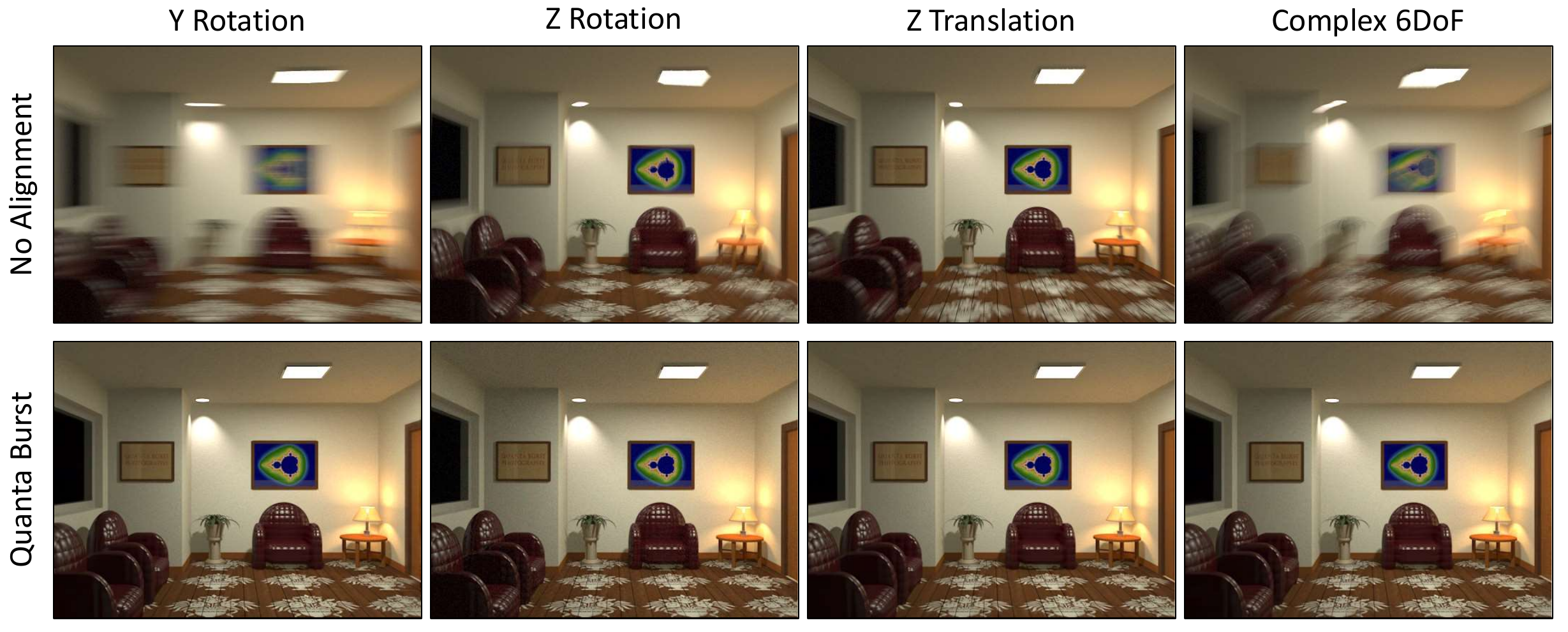}
  \vspace{-0.1in}\caption{\textbf{Performance for different types of camera motion.} We simulate four different types of motion for the same scene: rotation around y-axis, rotation around z-axis, translation along z-axis and a random 6 degrees-of-freedom (DoF) trajectory. In all cases, the proposed algorithm is able to align the binary images and generate high-quality images.}
  \vspace{-0.1in} \label{fig:sim_motion_type}
\end{figure*}

\begin{figure} 
  \includegraphics[width=\linewidth]{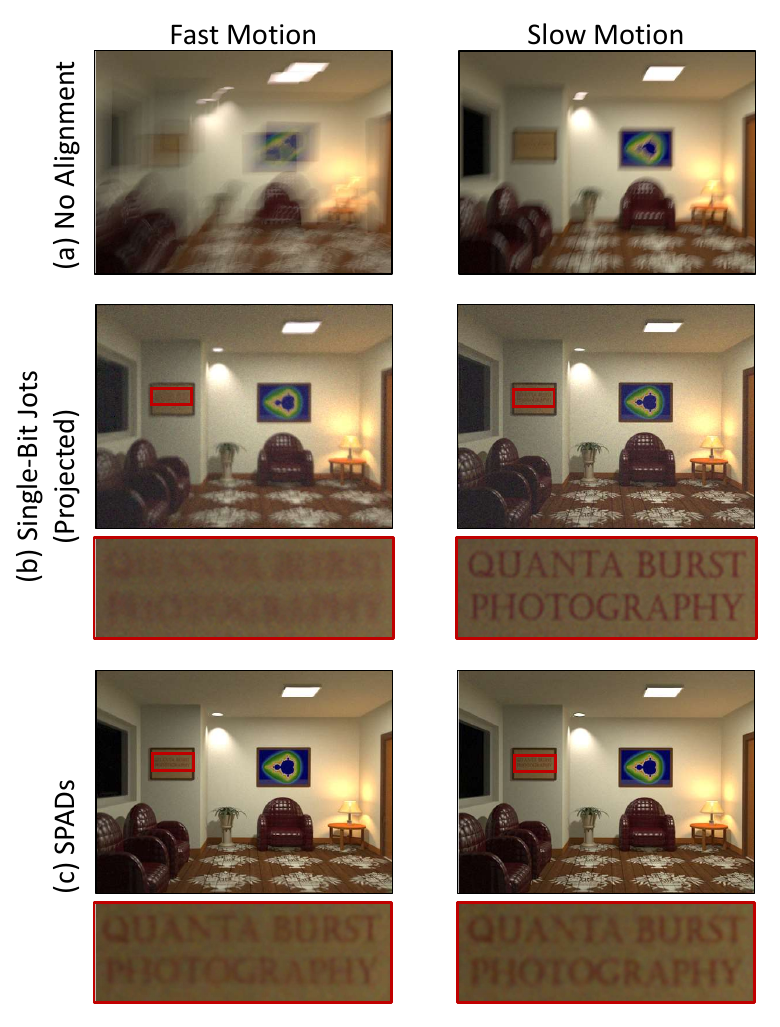}
 \vspace{-0.3in} \caption{\textbf{Comparison of jots and SPADs under different motion speeds.} We simulate a projected jots device with the same bandwidth as SPADs (and thus, higher spatial resolution). For fast motion, temporally super-sampled SPADs are able to resolve the motion blur and achieve sharper image. For slow motion, spatially super-sampled jots are able to reconstruct image details with higher fidelity.}
 \vspace{-0.1in} \label{fig:sim_jots_motion}
\end{figure}


\section{Results} \label{sec:results}
\subsection{Simulation Results} \label{subsec:sim_results}
\begin{table}%
\caption{Simulation Configuration}
\label{tab:sim_config}
\begin{minipage}{\columnwidth}
\vspace{-2mm}
\begin{center}
\begin{tabular}{l|ccc}
  \toprule
  Sensor Type  & Conventional & Jot & SPAD\\ \midrule
  Resolution     & \multicolumn{3}{c}{Same}\\
  Pixel Pitch & \multicolumn{3}{c}{Same}\\
  Bit Depth & 10 & 1 & 1\\
  QE / PDE (R) & 59\% & 64\% & 17\% \\
  QE / PDE (G) & 64\% & 71\% & 23\% \\
  QE / PDE (B) & 47\% & 62\% & 21\%\\
  Read Noise (per pixel) & 2.4$e^{-}$ & 0.24$e^{-}$  & 0 \\
  \begin{tabular}{@{}l@{}}Dark Current Noise / \\ Dark Count Rate (per pixel)\end{tabular} & 1$e^{-}$/s & 0.16$e^{-}$/s & 7.5cps\\
  \bottomrule
\end{tabular}
\end{center}

\end{minipage}
\vspace{0mm}
\end{table}%


We simulate the imaging process for a SPAD camera and a conventional camera of the same resolution and pixel pitch. We first simulate the ground-truth linear intensity images using a ray tracer (POV-Ray) and then draw Bernoulli samples according to Eq.~\ref{eq:bernoulli} to synthesize the binary images. Tab.~\ref{tab:sim_config} shows the sensor parameters we used for the simulation. The parameters for the conventional sensor are for a high-end machine-vision camera~\footnote{https://www.flir.com/products/grasshopper3-usb3/?model=GS3-U3-123S6C-C}. 
The parameters for the SPAD camera are based on the SwissSPAD2 sensor we use for our experiments.  Currently SwissSPAD2 does not have a Bayer filter for color imaging. We do not simulate the Bayer filter and demosaicing process but render the RGB channels directly. The alignment is performed on a grayscale version of the image and the merging is applied to the three channels independently. The fraction of incident photons that are measured by a SPAD is given by its photon detection efficiency (PDE), which is defined as the product of quantum efficiency, fill factor and photon detection probability (PDP). (See Sec.~\ref{sec:outlook} for a detailed discussion.) The PDE used in the simulation is computed by multiplying the PDP of SwissSPAD2 with the spectral response of a set of contrived color filters and the fill factor (assumed to be 50\% which can be achieved with microlenses~\cite{antolovic_optical-stack_2019}). The dark count rate is assumed to be spatially uniform (no hot pixels). For real images, this non-uniformity can be calibrated and compensated as shown in Eq.~\ref{eq:mle}.


\paragraph{Comparison of conventional and quanta burst photography.} 
We compare the results for single-shot conventional image, conventional burst photography and quanta burst photography for different lighting conditions. For conventional burst result, we use an approach similar to conventional burst photography methods~\cite{hasinoff_burst_2016}. The exposure time and number of bursts are determined using the strategy described in Sec.~\ref{sec:analysis}.


Fig.~\ref{fig:sim_lighting} shows the simulation results for different lighting conditions. The scene is kept static while the camera is moving. The trajectory of the camera is set to be a linear 3D translation plus a small, smooth random 6 degrees-of-freedom (DoF) perturbation at each frame. For a scene with sufficient light, both burst methods generate high-quality images. In low light, SPAD-based quanta burst photography generates much better result as there is no read noise.
Please refer to the supplementary report for a comparison of the two methods for different motion speeds under extremely low light. 

\paragraph{Performance for different types of camera motion.} Fig.~\ref{fig:sim_motion_type} \linebreak shows the quanta burst reconstruction results for different kinds of camera motion, including rotation around y-axis, rotation around z-axis, translation along z-axis and random 6DoF motion. In all cases, relatively blur-free images are reconstructed. 


\subsection{Comparison between Jots and SPADs} 
The quanta burst photography approach discussed so far is applicable to both single-photon sensing technologies: SPAD and jots. What are the relative benefits of the two technologies? In this section, we address this question by comparing their performance in various imaging scenarios. 


\paragraph{Adapting proposed approaches to spatially oversampling jots.} Due to the spatially oversampling nature of jots, the spatial resolution of raw jots images is typically higher than the final output image (oversampling factor $K>1$~\cite{yang_bits_2012}). A box filter is applied to downsample the raw binary images (related to the boxcar function used in \cite{chan_images_2016}) and convert them to floating point intensity values. The float images are then divided into temporal blocks as with SPADs (although with smaller block sizes than SPADs) and processed through the align and merge pipeline. 


\paragraph{Comparison under different amounts of motion.} Fig.~\ref{fig:sim_jots_motion} shows a comparison between the reconstruction results of SPADs and jots. We simulate two sequences of the same scene where the camera moves at different speeds. Since jots-based devices have yet to achieve a very high resolution (1024$\times$1024 so far), and their temporal resolution is lower than SPAD (1kHz vs 97.7kHz), we compare SPADs with a ``projected jot device'' with a resolution of $5120\times 5120$, such that total number of pixel measurements (data bandwidth) of the two sensors is the same. We assign the same data bandwidth to the two sensors based on the assumption that the bandwidth will be an important limiting factor for the frame rate for both sensors, as their specifications evolve in the future. 

Under fast motion, the merged image from jots contains motion blur, while SPADs are able to register the binary images and merge them into a sharp image. On the other hand, when the motion is slow, jots are able to generate a sharper image due to their high spatial resolution. Therefore, we envision these two technologies to complement each other: SPADs achieve higher performance in high-speed scenarios, while jots with projected high resolution will achieve better image quality for scenes with relatively slow motion and high-frequency texture details. The reader is referred to the supplementary technical report for more comparisons including those for multi-bit jots, and for comparisons using the sensor parameters of \emph{currently available} state-of-the-art prototypes for  jots~\cite{gnanasambandam_megapixel_2019} and SPADs~\cite{ulku_512_2019}.

\begin{figure} 
  \includegraphics[width=0.95\linewidth]{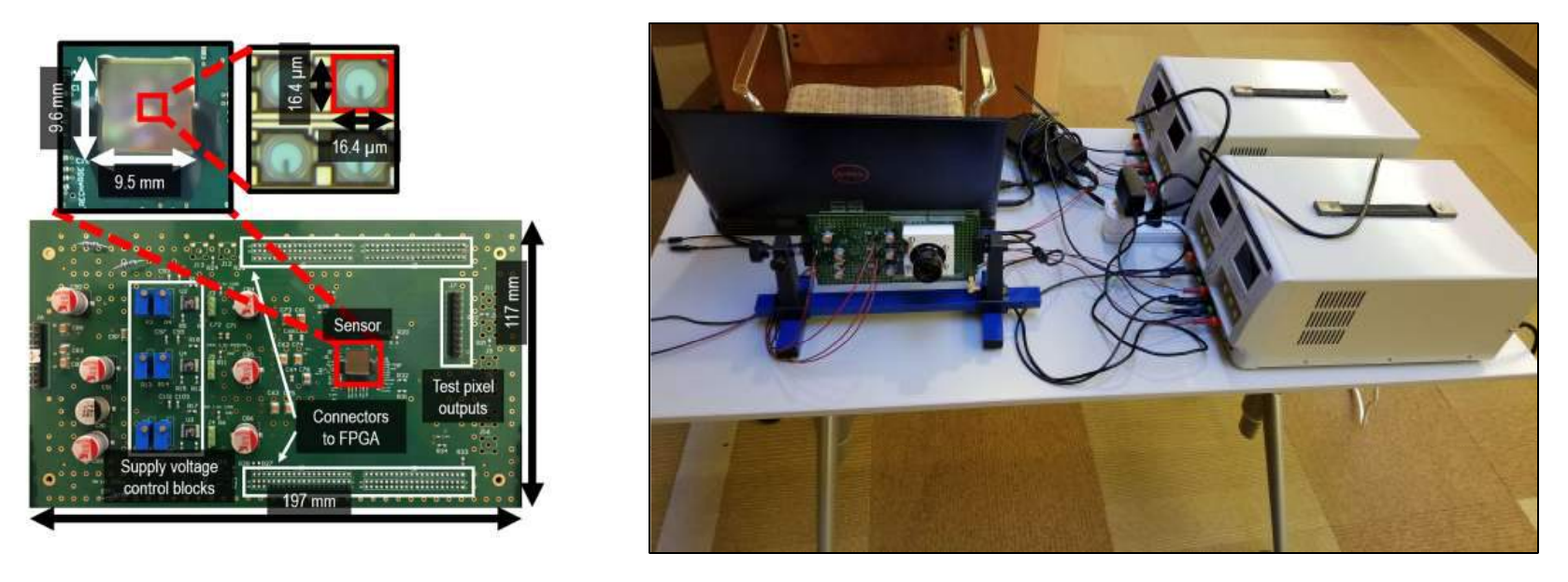}
  \vspace{-0.1in} \caption{\textbf{Camera setup.} \textbf{(Left)} The SwissSPAD2 board \cite{ulku_512_2019}. \textbf{(Right)} Camera setup.}
  \vspace{-0.0in}\label{fig:camera_setup}
\end{figure}

\begin{figure*} 
  \includegraphics[width=0.97\linewidth]{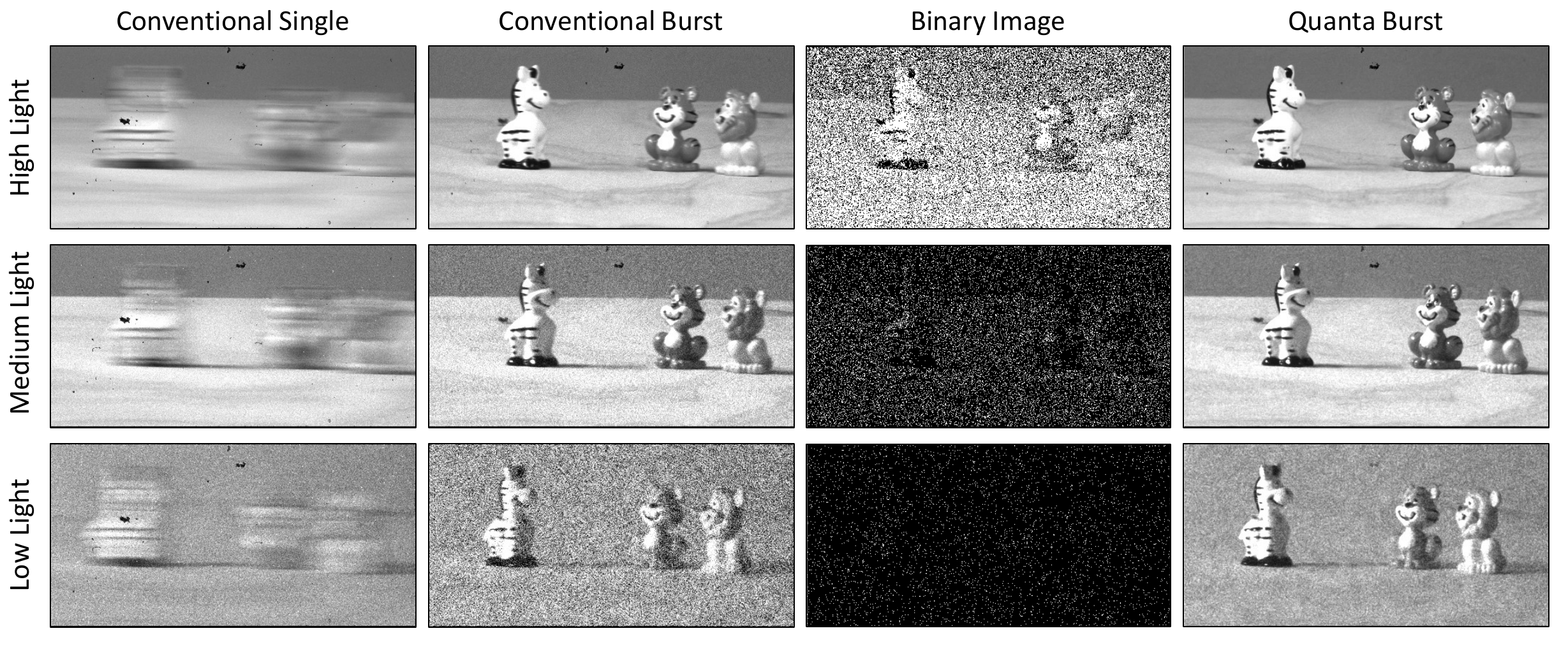}
 \vspace{-0.1in} \caption{\textbf{Performance under different lighting conditions.} We capture three 2000-frame binary sequences for the same scene under three different lighting conditions. A sample binary image from each sequence is shown in the third column. The binary images become sparser as the light level decreases. For conventional cameras, there is a trade-off between motion blur and noise, which makes it difficult to generate a high-quality image in low-light environments, either with a single long exposure (first column) or with a burst (second column). For quanta burst photography, it is possible to resolve fast motion without sacrificing the SNR (fourth column). Even in very low light, a reasonable image is reconstructed by aligning and merging the sparse and noisy binary frames.}
\vspace{-0.05in}  \label{fig:real_lighting}
\end{figure*}

\begin{figure*} 
\includegraphics[width=0.97\linewidth]{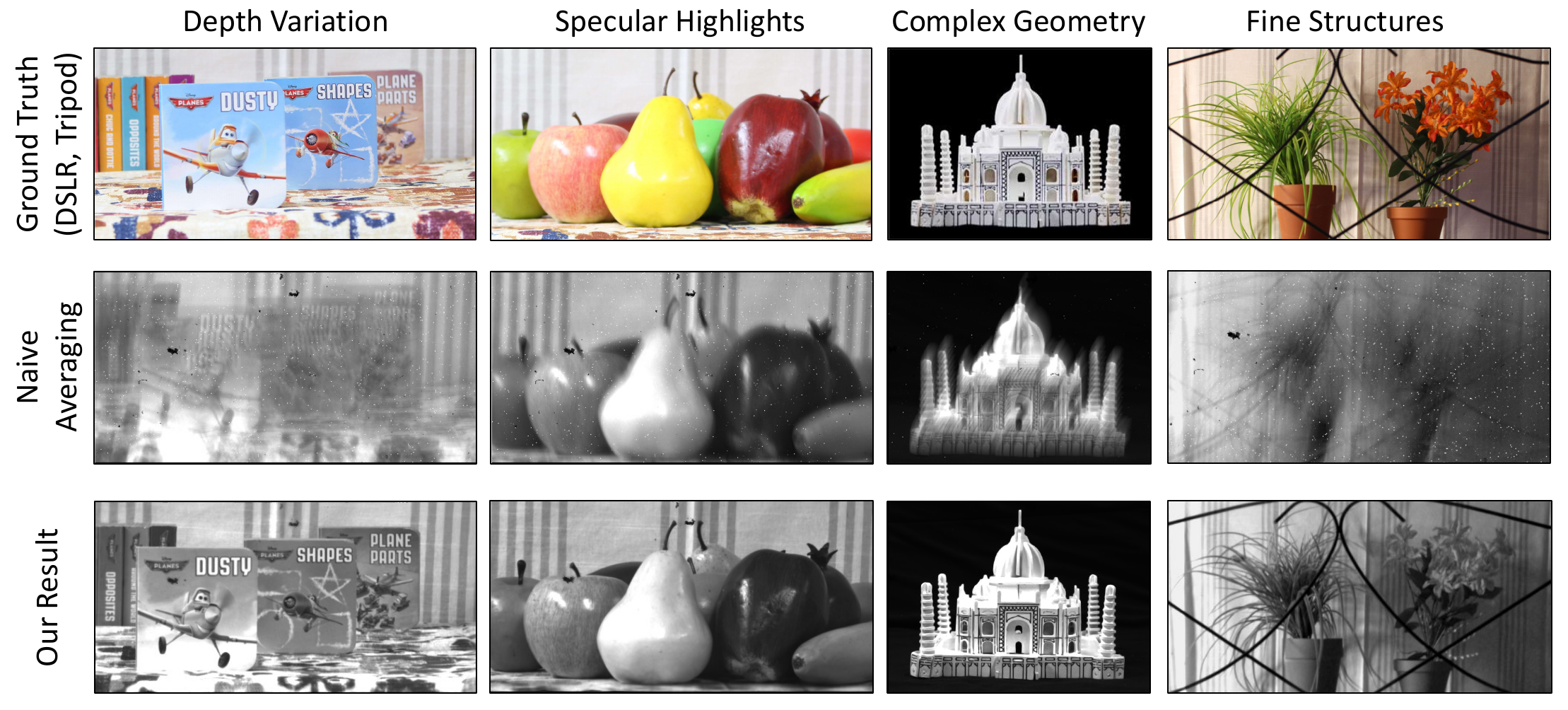}
  \vspace{-0.1in}  \caption{\textbf{Challenging scenes.} We show the reconstruction results of the proposed method for various challenging scenes involving high depth variation, specular highlights, complex scene geometry and fine structures. The camera was handheld, and follows a random 6DoF motion. Images are reconstructed from 10000 binary frames. In all cases, the proposed method is able to create a blur-free image with high SNR.}
\vspace{-0.15in}  \label{fig:real_challenge}
\end{figure*}


\begin{figure*} 
  \includegraphics[width=0.95\linewidth]{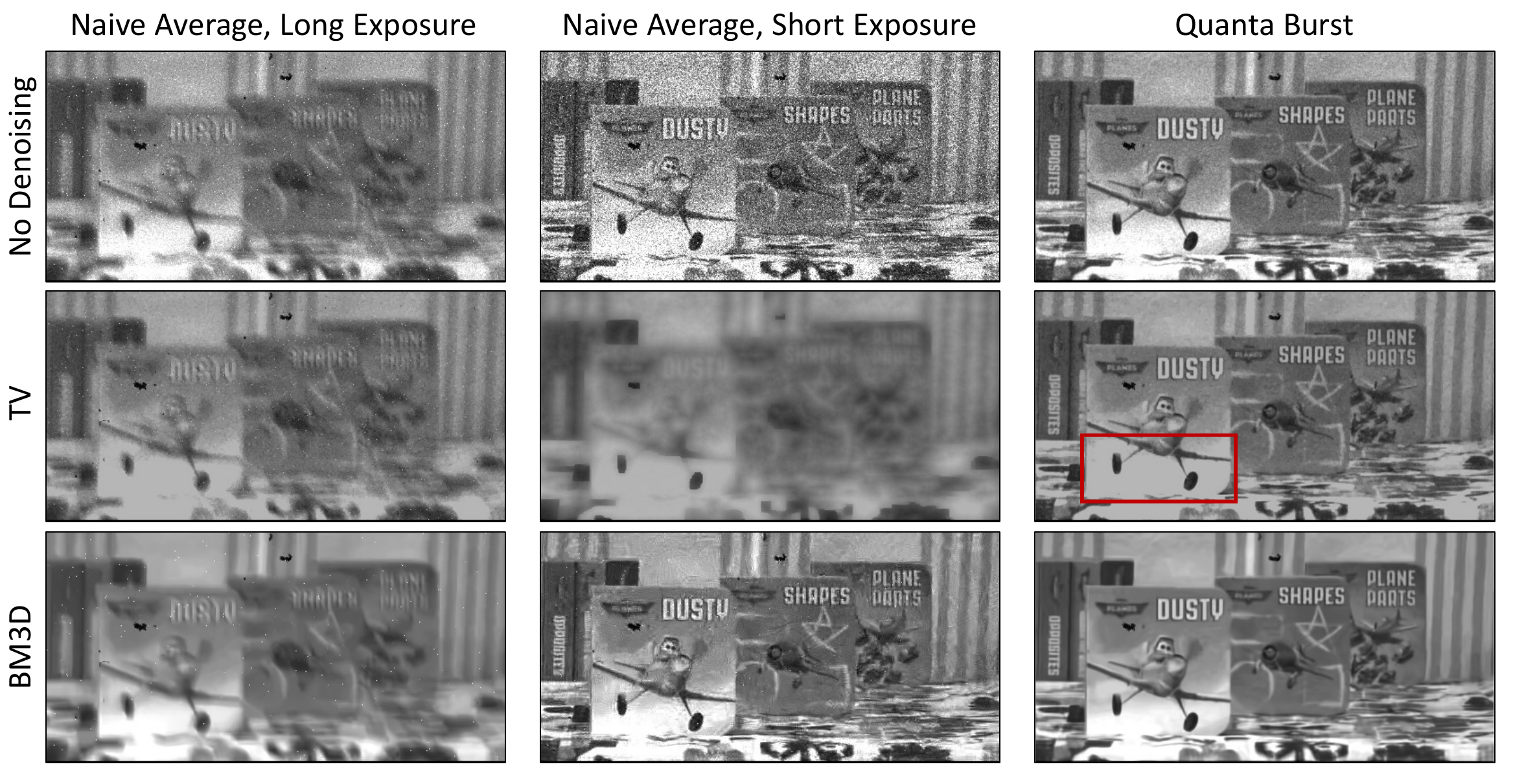}
  \vspace{-0.2in} \caption{\textbf{Comparison of denoising algorithms.} \textbf{(Left)} Naive average reconstruction without motion compensation on a long sequence (200 images). Results contain severe motion blur. \textbf{(Center)} Naive average reconstruction without motion compensation on a short sequence (20 images). Results are sharp but contain strong noise. Denoising algorithms reduce noise but also remove high-frequency image details. \textbf{(Right)} Burst align and merge results on 200 images. Results are sharp and less noisy. Applying denoising algorithms further reduces noise. BM3D outperforms TV, which results in oversmoothing for short exposure and loss of contrast for long exposure (red rectangle).}
  \vspace{-0.1in}\label{fig:exp_denoise_books}
\end{figure*}

\begin{figure*} 
  \includegraphics[width=0.95\linewidth]{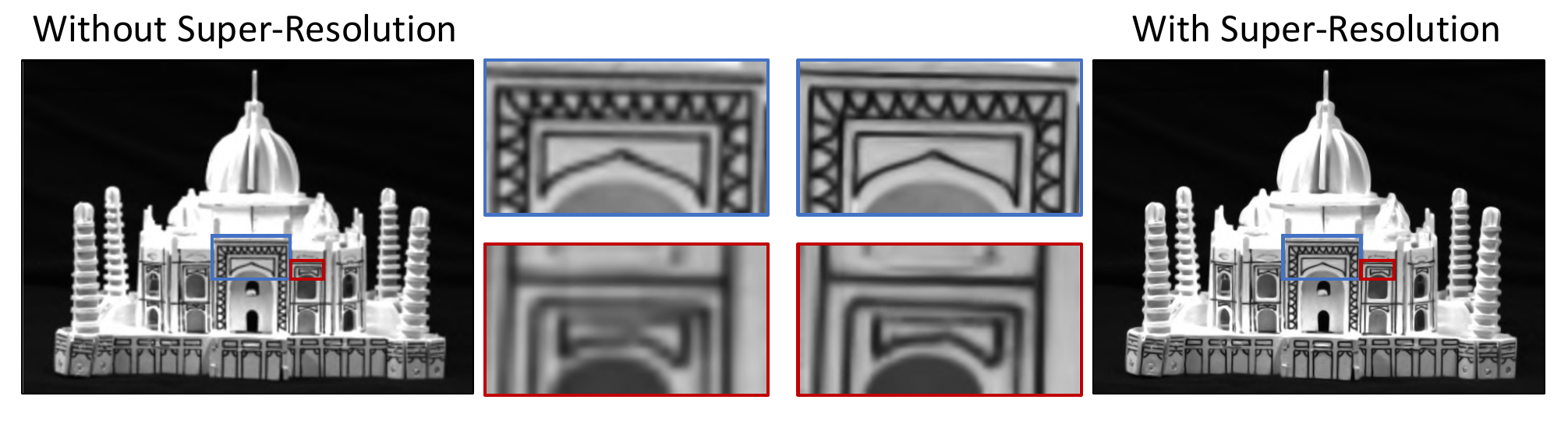}
\vspace{-0.2in}  \caption{\textbf{Achieving super-resolution.} We compare the output of the normal merging algorithm vs. the super-resolution algorithm. The super-resolution algorithm is able to reconstruct image at 2x resolution, creating sharper edges and mitigating aliasing artifacts.}
\vspace{-0.05in}  \label{fig:real_sr}
\end{figure*}

\begin{figure*} 
  \includegraphics[width=0.97\linewidth]{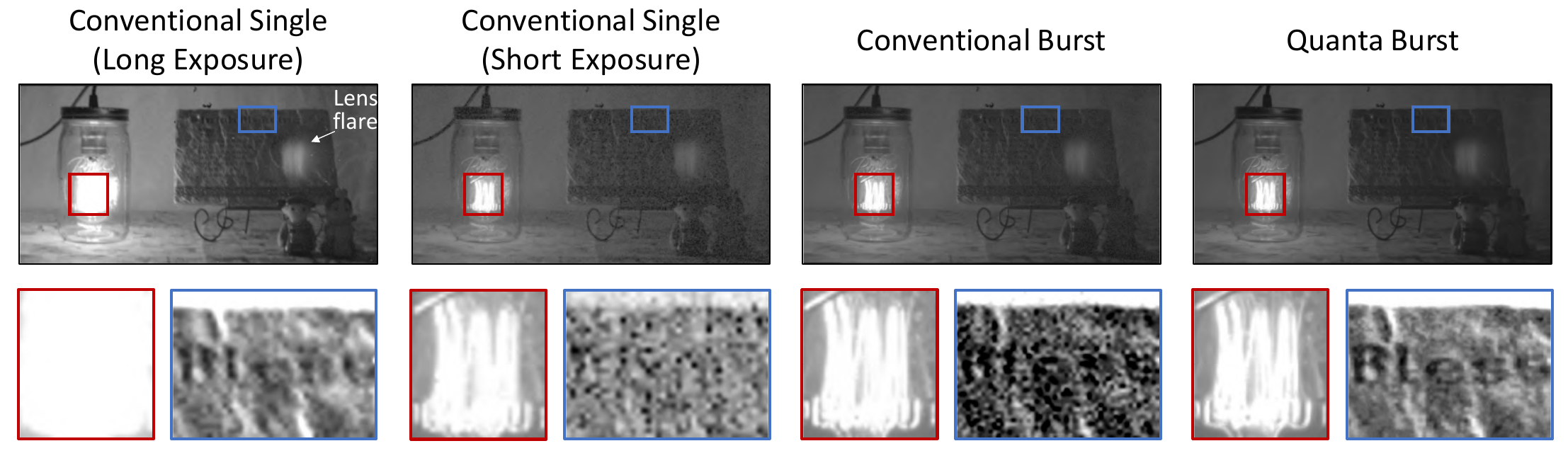}
  \vspace{-0.1in}  \caption{\textbf{Reconstructing high dynamic range scenes.} We capture a scene with high dynamic range where the light source (the lamp) is directly visible in the image. A single conventional image either gets saturated (long exposure) or fails to capture the details in the dark regions (short exposure). Conventional burst photography improves the dynamic range, but remains noisy in the dark regions due to read noise. Quanta burst photography achieves very high dynamic range and is able to recover the details of the filament and the text on the plaque at the same time. 100000 frames are captured to reconstruct the full dynamic range. All images are processed using the same tone-mapping algorithm~\cite{ashikhmin_tone_2002}.}
  \vspace{-0.00in}  \label{fig:real_hdr}
\end{figure*}

\begin{figure*} 
  \includegraphics[width=0.97\linewidth]{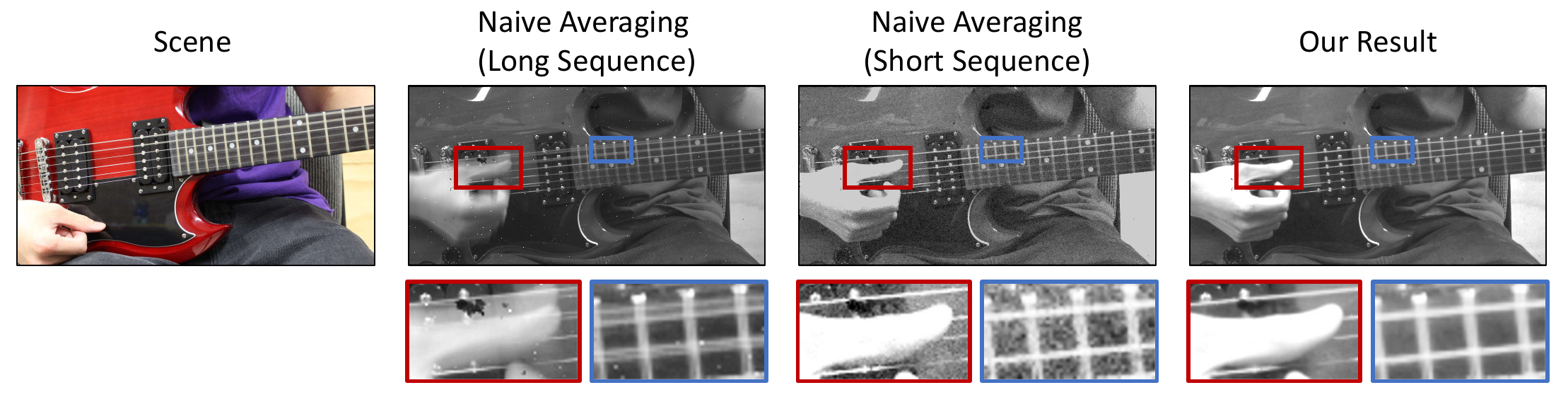}
  \vspace{-0.1in}  \caption{\textbf{Resolving scene motion.} A person plucking the lowest two strings of a guitar. Averaging the captured binary sequence results in either ghosting artifacts (long sequence with 2000 binary frames) or a low SNR (short sequence with 100 binary frames). Our method is able to reconstruct a high-quality image from 2000 frames despite fast and non-rigid scene motion.  {\bf See supplementary video for a short video reconstruction.}}
  \vspace{-0.00in}  \label{fig:real_scene_motion}
\end{figure*}

\begin{figure*} 
  \includegraphics[width=\linewidth]{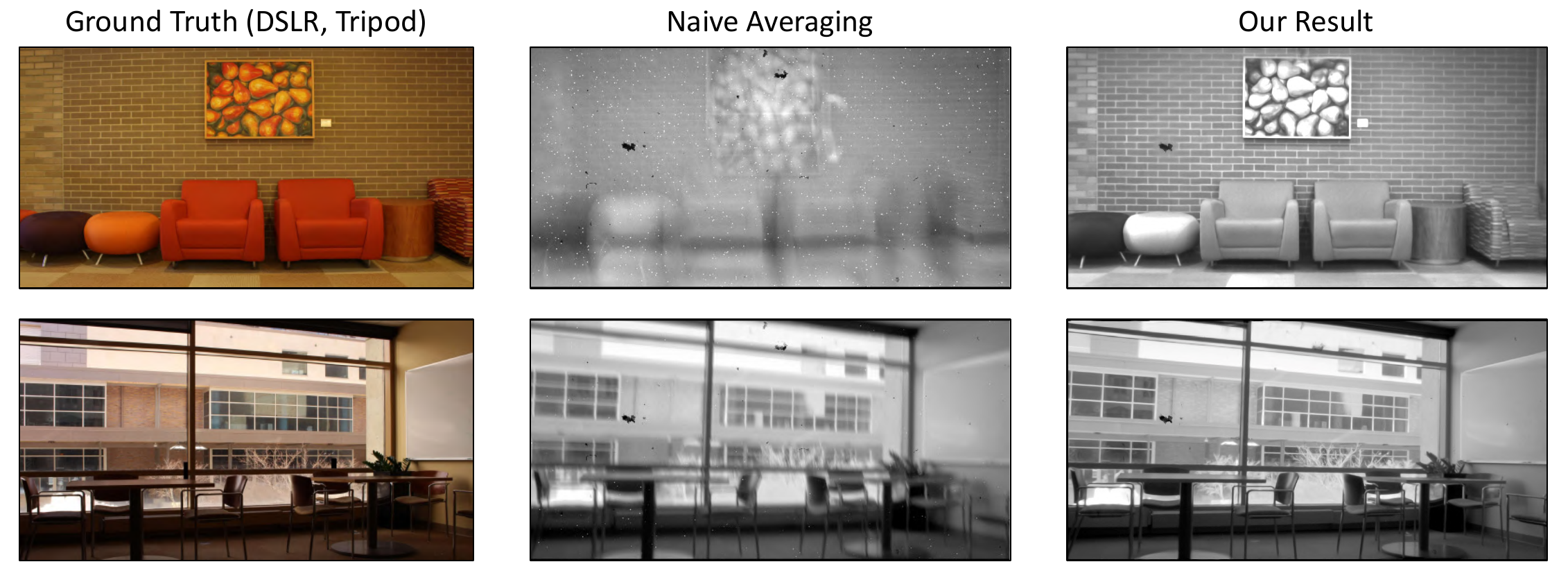}
  \vspace{-0.24in}  \caption{\textbf{Indoor scenes with different lighting.} The proposed method is able to recover sharp images despite the aggressive camera motion and high dynamic range, for various scenes under different, real-world lighting conditions.}
  \vspace{-0.0in}  \label{fig:real_indoor}
\end{figure*}

\subsection{Experiments}
We use a SwissSPAD2 camera~\cite{ulku_512_2019} to perform real experiments  (Fig.~\ref{fig:camera_setup}). This SPAD camera can capture binary frames at a spatial resolution of 512$\times$256. The maximum frame rate of the camera is 96.8kHz. The camera does not have microlenses and has a native fill factor of about 13\%. Currently the sensor is not equipped with Bayer filters, so only grayscale images are reconstructed. We identify the hot pixels by taking 100000 frames while covering the sensor completely from light sources. The hot pixels are corrected for each binary frame. See the supplementary report for details.

\paragraph{Performance for different lighting conditions.}  
Fig.~\ref{fig:real_lighting} shows the performance of quanta burst photography for different lighting conditions. We choose the same still scene for all sequences. The camera was moved horizontally to emsure the motion is controllable and reproducible across different sequences. The conventional camera images are emulated from the captured binary images by first reconstructing the intensity using Eq.~\ref{eq:mle} and then adding the read noise and quantization error according to the parameters in Tab.~\ref{tab:sim_config}. Quanta burst photography generates images with higher quality than conventional single and burst images. Even in very low light, where the individual binary frames are sufficiently sparse to make it nearly impossible to make out the scene structure, a reasonable image is reconstructed by aligning and merging the sparse and noisy binary frames. Please see the supplementary report for performance of the proposed method for different camera moving speeds.

The purpose of this experiment is not to compare a conventional sensor and SPAD sensor directly. In fact, due to the low resolution and low quantum efficiency of current SPAD sensors, the SPAD will almost always generate worse-quality images than a commercial CMOS sensor. Here we simulate the conventional images by assuming a conventional sensor with the same resolution and quantum efficiency as the SPAD array. Due to the blur-noise trade-off, conventional sensor struggles in reconstructing high-quality images, while SPAD has the potential of super-sampling in time and mitigate motion blur even for low-light and fast-moving scenes.

\paragraph{Reconstructing challenging scenes.} Fig.~\ref{fig:real_challenge} shows various scenes involving large depth variations, specular high lights, complex geometry and fine structures. Such scenes are usually challenging for optical flow and block matching algorithms. The camera was handheld, and underwent a random 6DoF motion when capturing the images. Since a long-focus lens is used, even natural hand tremor causes a large apparent motion in the image space. Despite these challenges, the proposed method is able to reconstruct blur-free images with high SNR.

\paragraph{Comparison of denoising algorithms.} After aligning and merging the binary frames into a sum image with low noise and blur, its SNR can be further improved via spatial denoising algorithms (\eg BM3D~\cite{dabov_image_2007}, total variation (TV)~\cite{chan_efficient_2014}). BM3D is applied as a post-processing step after the Anscombe transform, whereas total variation is formulated as a joint reconstruction and denoising optimization problem~\cite{chan_efficient_2014}. Fig.~\ref{fig:exp_denoise_books} compares the results of different combinations of spatio-temporal denoising schemes. Traditional single-photon image reconstruction (naive average) contains either motion blur in the long sequence, or heavy noise in the short sequence which cannot be perfectly removed using BM3D. In contrast, quanta burst photography approach in combination with spatial denoising is able to generate sharp, less noisy image. In our experiments, BM3D consistently performs better than TV, which results in over-smoothing for short exposure, and loss of contrast for long exposure. See the supplementary technical report for more comparisons.


\paragraph{Super-resolution.} Fig.~\ref{fig:real_sr} demonstrates the performance of the super-resolution algorithm. A high-resolution lens is used with the camera which creates aliasing in the image when the scene is perfectly in focus. The super-resolution algorithm is able to utilize the aliasing and sub-pixel motion between frames to create a higher-resolution image with sharper image details and less aliasing artifacts than the normal merging algorithm.

\paragraph{Reconstructing high dynamic range scenes.} Fig.~\ref{fig:real_hdr} shows a high dynamic range scene captured by the SPAD array. The only light source in the scene, the lamp (red box), is directly visible in the image, which is about 2000 times brighter than the text on the plaque (blue box), which does not receive any direct light. Similar as in Fig.~\ref{fig:real_lighting}, we simulate the conventional images by adding read noise and quantization error. With a single capture, the conventional image is either saturated around the lamp, or cannot recover the texts on the plaque. Conventional burst photography improves the dynamic range, but the text is still indiscernible due to read noise. Quanta burst photography is able to recover both the filament and the text at the same time.

\begin{figure*}
  \includegraphics[width=\linewidth]{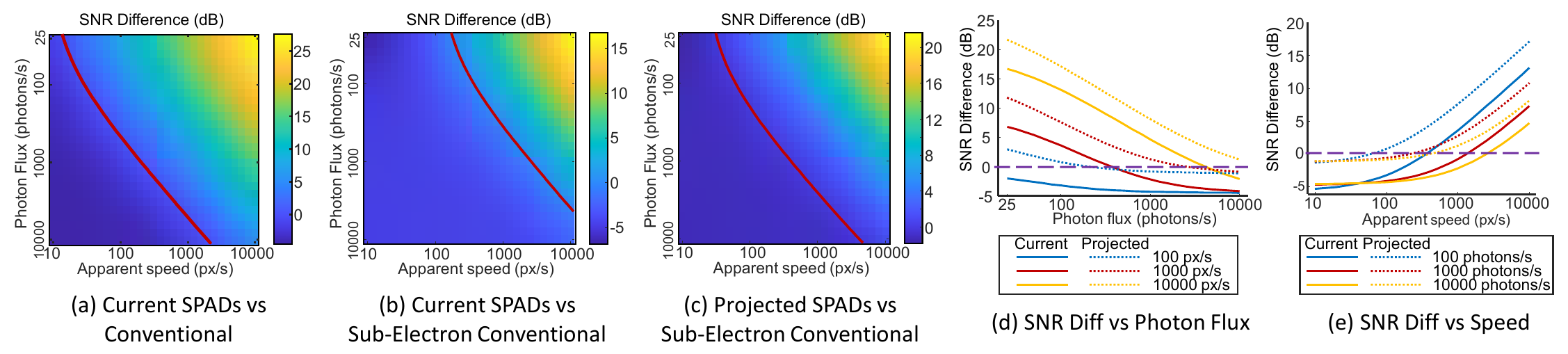}
  \vspace{-0.2in} \caption{\textbf{Theoretical SNR analysis.} \textbf{(a)} SNR difference between quanta burst photography based on current SPADs and conventional burst photography based on a machine vision CMOS sensor. ($SNR_{quanta}-SNR_{conv}$) in dB as a function of incident photon flux and apparent motion speed. SPADs achieve significantly higher SNR under very low light and high speed. On the other hand, in well-lit scenes with small motion, quanta burst photography performs worse due to lower quantum efficiency and higher dark current noise. The red line indicates the iso-contour for SNR difference = 0 (equal performance). \textbf{(b)} SNR difference between current SPADs and the recent conventional image sensor on iPhone 7. The conventional sensor works better for a wider range of flux intensity and apparent speeds due to its sub-electron read noise. \textbf{(c)} SNR difference between projected SPADs with PDE = 50\% and iPhone 7 sensor. \textbf{(d, e)} 1D slices of the 2D plots in (b) and (c) by fixing a specific flux or speed. In each case, the difference is higher for low light levels and large motions. } 
  \vspace{-0.1in}\label{fig:snr_analysis}
\end{figure*}

\paragraph{Resolving scene motion.} Since the proposed method only computes patch-wise motion and does not assume any global motion model, it is capable of resolving scene motion. Fig.~\ref{fig:real_scene_motion} shows a person plucking the lowest two strings on the guitar. Simple averaging of binary frames creates ghosting artifacts or strong noise. Our method is able to resolve the plucking motion of the thumb and the vibration of the strings with lower noise.

\paragraph{Indoor scenes with different, natural lighting.} In addition to the controlled scenes in the lab, we captured a few indoor scenes with more natural lighting. As shown in Fig.~\ref{fig:real_indoor},  the proposed method is able to reconstruct high-quality images under these unstructured environments. Please refer to the supplementary technical report for more simulation and experimental results.

\section{When to Use Quanta Burst Photography?} \label{sec:analysis}
What are the imaging regimes where quanta burst photography can outperform conventional cameras?~\footnote{This analysis is not meant to be a direct comparison between \emph{current} single-photon and conventional cameras. Conventional CMOS sensors have considerably higher spatial resolution and color filters, and thus, will achieve better image quality as compared to current SPAD arrays in the foreseeable future. The goal of this analysis is to provide guidelines on when using quanta burst photography can be beneficial, assuming SPAD arrays can match the spatial resolution of sCMOS sensors. } To address this question, we characterize the performance of conventional and quanta burst photography in terms of the SNR of the reconstructed linear image: 
\begin{equation}
    SNR = 20\log_{10}\frac{\hat{\phi}}{\rmse(\hat{\phi})}\,
\end{equation}
where $\hat{\phi}$ is the estimated image intensity, and $\rmse{\hat{\phi}}$ is the root mean squared error of the estimate. We assume that the input images are perfectly aligned (no mis-alignment errors) for both conventional and single-photon cameras, so that the estimation error is only due to image noise.

\paragraph{Conventional cameras:} The image formation of conventional image sensors is given by an affine model~\cite{hasinoff_noise-optimal_2010}:
\begin{equation}
    I = Z + \epsilon_{rc} + \epsilon_{dc}\,,
\end{equation}
where $Z\sim Pois(\phi\tau_c\eta_c)$ is the photon counts as in Eq.~\ref{eq:poisson} ($\tau_c$ and $\eta_c$ are the exposure time and quantum efficiency for the conventional sensor). $\epsilon_{rc}\sim  N(0, \sigma_{rc})$ is the read noise. $\epsilon_{dc}\sim Pois(\tau_c r_c)$ is the dark current noise caused by thermal current with flux $r_c$. These three components are statistically independent of each other. To simplify the analysis, we assume all images are captured at the same ISO speed and temperature such that $\sigma_{rc}$ and $r_c$ are fixed. 

Suppose a conventional burst photography algorithm captures a burst of $n_c$ images. The process of merging the captured images into a result image can be viewed as a maximum likelihood estimation process. Assuming the images are perfectly aligned such that the $n_c$ images can be merged simply by taking their average:
\begin{equation}
    \hat{\phi}_c = \frac{1}{n_c\tau_c\eta_c}\sum_{t=1}^{n_c}(I_t - \tau_c r_c)\,,
\end{equation}
where $I_t$ is the image captured at time $t$. We assume the dark current noise can be calibrated at each pixel. The mean of the calibrated dark current noise is subtracted from the sum of images to give an unbiased estimate of the photon flux (linear intensity image).


From the noise model, the root mean squared error (RMSE) of this estimator due to noise variance is given by
\begin{align} \label{eq:rmse_c}
    \rmse(\hat{\phi}_c) = \sqrt{\Var[\hat{\phi}_c]}&=\sqrt{\frac{\phi\eta_c+r_c}{T\eta_c^2}+\frac{n_c\sigma_{rc}^2}{T^2\eta_c^2}}\,,
\end{align}
where $T=n_c\tau_c$ is the total exposure time for the sequence.


\paragraph{SPAD cameras:} A maximum likelihood estimator for SPAD camera is derived in Eq.~\ref{eq:mle}. For a sufficiently long sequence $n_q>30$, the variance of the MLE can be estimated using Fisher information (See the supplementary technical report for the derivation):
\begin{equation}\label{eq:rmse_q}
    \rmse(\hat{\phi}_q) = \sqrt{\Var[\hat{\phi}_q]} \approx \frac{1}{\sqrt{I({\phi})}} = \sqrt{\frac{e^{\phi\tau_q\eta_q+r_q\tau_q}-1}{n_q \tau_q^2 \eta_q^2}}\,,
\end{equation}
where $\tau_q$ and $\eta_q$ are the exposure time and quantum efficiency for the single-photon camera. 

The RMSE for both modalities depend on the total exposure time $T$ of the image sequence (assumed same for both modalities for a fair comparison) and the total number of frames $n_c$ and $n_q$, which, in practice, in turn depend on the photon flux level $\phi$ and camera motion: longer exposure is preferred when the light level is low and the camera is moving slowly. \cite{liba_handheld_2019} proposes ``motion metering'' which automatically selects the exposure time based on a prediction of future scene and camera motion. We take a similar approach for our analysis: we assume the scene and camera motion are known or can be estimated such that $T$ and $n$ can be determined according to the following three principles: (1) When the motion is slow, the total exposure time is chosen to meet a target total number of photons to ensure high SNR. (2) When the motion is fast, the total exposure time is limited by a maximum amount of motion across the sequence. (3) The total number of frames is chosen to ensure the per-frame motion blur is below a threshold. Details about the strategy can be found in the supplementary  report. The SNR of both approaches can then be expressed as a function of photon flux and camera motion, which allows comparison of the two approaches.


\paragraph{SNR comparisons between conventional and SPAD cameras:} Fig.~\ref{fig:snr_analysis} plots the difference of SNRs ($SNR_{quanta}-SNR_{conv}$) in dB for a wide range of photon fluxes and apparent speeds. Fig.~\ref{fig:snr_analysis} (a) compares the burst photography performance between a current SPAD sensor and a machine-vision CMOS sensor with parameters listed in Tab.~\ref{tab:sim_config}.  At ultra low light and high speeds, the SPAD sensor performs considerably better than the CMOS sensor (up to 27.5dB = 23.7 times). On the other hand, in well-lit scenes with negligible motion, the SPAD performs worse (albeit at most by a factor of $0.5$) due to relatively low PDE and high DCR of current SPAD arrays. 

Recently, advanced CMOS sensors used in high-end cellphones have achieved sub-electron read noise. Fig.~\ref{fig:snr_analysis} (b) plots the SNR difference between current SPADs and iPhone 7's sensor, which is reported to have a read noise of 0.68 electrons~\cite{claff_input-referred_nodate}. Such low read noise makes its performance better than current SPADs for a wider range of flux intensity and motion speeds. Since SPADs are an emerging technology, their specifications (in particular, resolution and PDE) continue to improve, arguably at a faster rate than conventional sensors which are already a mature technology. Fig.~Supp\ref{fig:snr_analysis} (c) compares iPhone 7's sensor with a projected SPADs which achieve a PDE of 50\%. To visualize the variations of the SNR difference with respect to one specific parameter, we show 1-D slices of the comparison between iPhone 7 and current/projected SPAD sensor in (d) and (e), where either the photon flux or the apparent speed is fixed. These figures demonstrate how the proposed analysis framework can be used to direct future development of SPADs for best performance under certain light levels and amount of motion. A theoretical dynamic range analysis can be found in the supplementary technical report.

\section{Outlook on Single-Photon Sensors} \label{sec:outlook}

In this section, we discuss the current state and future outlook of SPAD sensor arrays, in terms of their key characteristics: spatial resolution, temporal frame rate, photon detection efficiency (PDE), and the dark count rate (DCR).

\paragraph{Spatial resolution:} Due to their compatibility with mainstream CMOS fabrication lines, it was predicted in 2008 that SPAD image sensors could reach large resolutions within one decade~\cite{Charbon:2005,Charbon:2008}. In recent years, significant effort has been devoted to achieve this goal, with the world's first 1 MPixel SPAD array reported recently~\cite{Morimoto:2019}. With the same fabrication process, it is possible to go up to 5-10 MPixel, not far from their counterparts in CMOS imagers in several cell-phone cameras. Can we go even higher (e.g., 50 MPixel) in the long term? The key factor that limits the spatial resolution is the minimum pixel pitch, which in turn is limited by the necessity of placing a \emph{guard ring}~\footnote{A SPAD pixel detects single photons by creating an avalanche of photo-electrons (large current) when a photon is incident, and sensing the avalanche current via a comparator or a high-gain amplifier. A guard ring is a region around each SPAD pixel that forces the avalanche to be confined in the region, in order to prevent edge breakdown. Guard rings are implemented via geometric structures that are not sensitive to light.} around each SPAD pixel. In current CMOS technologies, due to the guard ring, SPAD pitch cannot be reduced below $1\mu $m. At that pitch, the guard ring occupies a large portion of the pixel, thus reducing the fill factor to a minimum. This limitation could be addressed via 3D-stacking~\cite{pavia_1_2015}, a potentially effective way to reduce SPAD pixel pitch by moving all the active and passive components associated with a SPAD pixel to the bottom tier of the sensor.


\paragraph{Frame rate and power consumption:} The frame rate of a SPAD sensor array is limited by the bit-rate the chip can deliver and by the number of communication channels it can host. For example, a 1 Mpixel camera with a frame rate of 1kfps, will generate 1Gbps of data, which can be handled by a single LVDS (low-voltage differential signalling) communication channel. Typically, this kind of channel requires about 10mW of power at full speed. If one wants to increase the frame rate by, say, $100 X$, then the data rate will increase to 100Gbps, with $1$W of power required, which may be prohibitive for consumer devices. This assumes that the internal power dissipation due to SPADs and chip operation is negligible, and that the readout speed of the pixels internally is not the bottleneck. The communication power consumption can be mitigated by performing on-chip image processing operations, and designing more efficient motion computation and image alignment operations that are amenable to on-chip processing. Furthermore, it is possible to exploit the spatio-temporal sparsity in the photon-cube raw data in low-light scenarios. Depending on the light-level in the scene, one could achieve a considerable data rate reduction by compressing the raw photon-cube data~\cite{Zhang:2018}.

\paragraph{Photon detection efficiency (PDE):} PDE is defined as the product of the pixel fill factor, and the photon detection probability (PDP), which is the probability that an impinging photon generates a detectable signal. PDP is the product of quantum efficiency and the probability of triggering an avalanche. PDP is dependent on the wavelength of photons; for current devices, the PDP is typically $50-70\%$ at $450-550$ nm. Due to low fill factors, earlier SPAD arrays had PDEs as low as $1 \%$ making them highly inefficient due to significant light loss. However, the PDE in recent arrays has increased to approximately $40 \%$ by using microlens arrays, which increase PDE by effectively increasing the fill factor. While still lagging the quantum efficiency of conventional sensors (approximately $60-90 \%$), the PDE of SPAD arrays will likely improve due to improving fabrication processes, including 3D stacking.

\paragraph{Dark count rate (DCR):} DCR is the rate of avalanche counts unrelated to photons, measured in counts-per-second (cps). Earlier SPAD devices were largely considered impractical due to high DCR, up to several tens of cps at cryogenic temperatures, and tens of kcps at room temperature. 
Fortunately, for current devices, DCR has been drastically reduced to $2$ cps~\cite{Morimoto:2019}, even at room temperature. Since SPADs do not have read noise, this DCR is sufficiently low to achieve nearly shot-noise-limited SNR, even in ultra low-light. Since DCR is proportional to the active area of a SPAD, as the pixels become smaller, DCR could be further reduced.

\section{Limitations and Discussion}

\paragraph{Resolving larger range of motions.}
The proposed alignment algorithm assumes the motion of the spatial image patches can be approximated by a 2D translation, which is usually appropriate for camera motion and rigid object motion. When this assumption does not hold, the discrepancies between the true deformation of the patch and the translation approximation can be mitigated by the robust merging algorithm. However, when the scene contains several small objects or undergoes nonrigid motion, such an approximation no longer holds, which can result in artifacts in the merged image. An interesting future research direction is to design optical flow algorithms for aligning images for such challenging scenes.

\paragraph{Fast, energy-efficient processing.}
Currently, our algorithms are implemented in unoptimized MATLAB code which takes about 30 minutes for processing a sequence with 10000 binary frames, which is far from real-time. For consumer photography applications, it is critical to perform the processing in a fast and also energy-efficient way. In our current implementation, the binary frames are treated as real numbers (\eg when warping them during the merging stage). One potential way to improve the efficiency is to utilize specialized computing architectures and algorithms for binary data~\cite{daruwalla_bitbench_2019,pfeiffer_deep_2018}.

\paragraph{Bandwidth limitation.} The high dynamic range of SPADs comes at the cost of large bandwidth requirement.  Currently, the captured binary images are stored on-board, and then transferred to a PC and processed offline. The bandwidth requirement can be relaxed by capturing multi-bit images (which sacrifices temporal resolution). The bandwidth in future SPAD sensors can also be improved by using faster interfaces such as PCIexpress and CameraLink.


\paragraph{Video reconstruction.} The proposed method can be used for reconstructing videos by shifting the reference frame in time. {\bf An example reconstructed video for the guitar sequence is shown in the supplementary video.} While the current approach reconstructs the video sequence one frame at a time, novel algorithms that enforce temporal coherency across reconstructed frames could be developed, resulting in improved video quality as well as lower computational complexity.


\paragraph{Free-running SPADs.} The proposed techniques are designed for SPAD arrays operating in synchronous clock-driven mode where all the pixels read off measurements simultaneously, at fixed intervals. It has recently been shown that event-driven~\cite{antolovic_dynamic_2018} or free-running SPADs~\cite{ingle_high_2019} achieve a higher dynamic range by recharging the SPAD as soon as the dead time due to a photon detection is over. An interesting future direction is to design quanta burst photography techniques for asynchronous SPAD arrays where pixels return binary measurement independently.

\paragraph{Quanta image processing pipeline.} The primary focus of this paper is on the alignment and merging of binary images. We apply denoising and tone-mapping as a post-processing step to the merged images. For modern camera systems with color filter arrays, there are several other essential processing steps in the image processing pipeline including demosaicking, white balancing, and dehazing. Specifically, demosaicking is a non-trivial problem since temporal interpolation of alignment can introduce color artifacts. Recent research suggests that there is a potential benefit of performing end-to-end processing from the raw sensor data~\cite{heide_flexisp_2014,gharbi_deep_2016,chen_seeing_2019}. A promising next step is to design a similar framework for quanta burst photography and explore whether similar benefits exist for single-photon images.

\begin{acks}
This research is supported in part by the \grantsponsor{GS:DARPA}{DARPA REVEAL program}{}, a \grantsponsor{GS:WARF}{Wisconsin Alumni Research Foundation (WARF) Fall Competition award}{} (UW-Madison), the \grantsponsor{GS:SWISS}{Swiss National Science Foundation Grant}{}~\grantnum{GS:SWISS166289}{166289} and \grantsponsor{GS:NE}{The Netherlands Organization for Scientific Research Project}{}~\grantnum{GS:NE13916}{13916} (EPFL). 
\end{acks}

\bibliographystyle{ACM-Reference-Format}
\bibliography{main, main_Mohit}

\end{document}


\maketitle

In this report, we provide additional technical details, analysis and experimental results that are not included in the main paper for better presentation. 

\section{Technical Details for the Proposed Algorithm}
\subsection{Removal of Hot Pixels}
Dark count rate (DCR) distribution on a real SPAD array is usually non-uniform in space. A few pixels which have very high DCR will always show up as bright pixels in the block-sum image, which are usually called ``hot pixels''. Such pixels can be identified by taking a dark image and locate the pixels with high counts. 

To remove them in the reconstructed image, one naive method is to perform a median filtering after merging. However, the existence of hot pixels interferes with the aligning process, especially in very dark scenes. Since hot pixels have very high intensity and do not move as the camera or scene moves, they bias the motion estimate towards zero motion, causing systematic alignment error. 

Another potential approach is to exclude the hot pixels in the data terms during alignment. This results in correct alignment, however when the binary images are warped in the merging stage, the hot pixels will sweep along the estimated motion trajectory, causing a ``hot stripe'' in the merged image.

Following the analysis above, it is necessary to correct the hot pixels in the binary frames. We adopt a simple approach by randomly assigning a hot pixel the binary value of one of its spatial neighbors. This approach essentially applies a spatially averaging filter at the binary frame level, which has shown to remove most of the hot pixels in real experiments.

\subsection{Choice of Block Size for Motion Estimation} 
Block size is an important parameter for achieving accurate frame alignment. If the block size is too small, each block has a low photon count (low SNR), which results in high alignment error. On the other hand, if the block size is too large, the block alignment is computed only at sparse timestamps, which is unable to capture high-frequency variations in the camera or scene motion. The optimal block size for a specific application depends on both the light level and the motion variations, and can be determined from prior knowledge of approximate scene flux levels and motion amounts. An interesting extension is to automatically determine the block size by performing light and motion metering. In the experiments, we use block sizes ranging from 100 to 500 frames.

\subsection{Choice of Block Size for Robust Merging} The block size used in the merging stage does not have to be the same as in the aligning stage. Once we estimate the fine-scale inter-frame motion field, the binary frames can be warped separately and grouped into blocks using a different block size. The choice of block size plays an important role in the merging stage as well. Too small a block size preserves the shot noise. Extremely large block sizes preserve alignment artifacts, since the frames within a block are simply added together. The optimal choice of block size again depends on the light level and the variation in scene/camera motion.

\subsection{Choice of Patch Size}
Patch size is another important factor on the image quality. Choosing a large patch size is likely to gather more photons and capture more features of the scene, which makes alignment easier. However, large patch sizes cannot correctly model nonrigid scene motion, resulting in motion artifacts. Choice of block size is especially important for current SPAD sensors whose spatial resolution is relatively low. We use $16x16$ patches for most experiments, $8x8$ for nonrigid scene motion, and $32x32$ for dark scenes with global motion.

\subsection{Super-Resolution}
In this section we give details on the super-resolution algorithm.
As mentioned in the main paper, after computing the fine-scale motion field, the binary frames are combined into warped block-sum images and filtered using the Wiener filter:
\begin{equation} \label{eq:wiener_sr}
    \hat{S}^w_f(\omega)=\hat{S}^w_{ref}(\omega)+A_i(\omega)(\hat{S}^w_{aux,i}(\omega)-\hat{S}^w_{ref}(\omega))\,.
\end{equation}
This step is used to reduce the misalignment artifacts, which plays a similar role as the point-wise robustness factor in \cite{wronski_handheld_2019}. As mentioned in the main paper, we found this step more robust, at the cost of computational complexity.

Instead of summing them up to form a merged patch in the original-resolution grid as in the original merging algorithm, 
each patch is treated as a bag of samples and warped to a higher-resolution grid. The value at each pixel of the high-resolution grid is computed by combining all samples in a neighborhood using a anisotropic kernel:
\begin{equation}
    S_{SR}(x,y) = \frac{\sum_{i\in \mathscr{N}}w_i\cdot S_i}{\sum_{i\in \mathscr{N}}w_i}\,,
\end{equation}
where $\mathscr{N}$ is the set of all sample points in the neighborhood around pixel $(x,y)$. $S_i$ is the photon counts of the $i$-th sample point. $w_i$ is the weight given by the anisotropic Gaussian kernel,
\begin{equation}
    w_i = \exp\left(-\frac{1}{2}(\vec{x_i}-\vec{x})^T\mat{\Omega}^{-1}(\vec{x_i}-\vec{x})\right)\,,
\end{equation}
where $\vec{x}=(x,y)$ is the pixel location of interest, $\vec{x_i}=(x_i,y_i)$ is the location of the sample point. 
The shape and size of the anisotropic kernel (encoded in the covariance matrix $\mat{\Omega}$) is determined by the analysis of the local structure tensor of a guide image. The guide image can be either the reference block-sum image (for faster computation) or a original-resolution reconstructed image obtained by running the normal merging algorithm beforehand (for better quality). For a flat region, a larger kernel is used to gather more pixels for denoising. For an edge, the kernel is stretched along the edge to avoiding over-smoothing and mitigate the alignment error around the edges. For a corner or local window with high variations, a small kernel is chosen to preserve the details (Fig.~\mainFigSr in the main paper).

The exact kernel design is very similar to the heuristics used in \cite{wronski_handheld_2019}. The only difference is that we set an upper bound to the anisotropy factor $A$ which determines the ratio of the major axis and short axis of the elliptical kernel:
\begin{equation}
    A = 1 + \min(\sqrt{\lambda_1/\lambda_2}, 5)\,
\end{equation}
where $\lambda_1, \lambda_2$ are the eigenvalues of the local structure tensor. This prevents the kernel from begin elongated too much, which causes artifacts along the edges. The exact parameters we use for the images we capture with SwissSPAD2 cameras are: $T_s=[8,16]$ depending on light level, $k_{detail}=0.3,k_{denoise}=1,D_{th}=0.005,D_{tr}=0.5,k_{stretch}=1,k_{shrink}=1$.


\smallskip\noindent\textbf{Using small block sizes.} 
Compared to original-resolution merging, super-resolution merging requires a larger number of blocks (smaller block size). This is because super-resolution benefits from larger number of sample points with different sub-pixel offsets, which grows with the number of blocks. As a result, super-resolution merging may suffer from the noise problem due to small block size (see Sec.~\mainSecMerge in the main paper). Our approach to solve this problem is to replace the reference block image $S_{ref}$ in Eq.~\ref{eq:wiener_sr} with the pre-reconstructed guide image, which has a much higher SNR. The noise estimate $\sigma$ and the scaling factor $c$ need to be adjusted correspondingly.

\section {Further Discussion on When to Use Quanta Burst Photography} \label{sec:analysis}
\subsection{Derivation of Eq.~\mainEqRmseQ}
Here we give a derivation of Eq.~\mainEqRmseQ in the main paper which gives the RMSE of the maximum likelihood estimator for quanta burst photography.
Recall that the binary value $B$ at a SPAD pixel follows a Bernoulli distribution:
\begin{equation}
    \begin{aligned}
        P\{B=0\} &= e^{-(\phi\tau\eta+r_q\tau)} \,,  \\
        P\{B=1\} &= 1-e^{-(\phi\tau\eta+r_q\tau)} \,,
    \end{aligned}
\end{equation}
where $\phi$ is the photon flux incident at the pixel, $\tau$ is the exposure time, $\eta$ is the quantum efficiency, $r_q$ is the dark count rate. 

The sum image is defined as the sum of all binary images
\begin{equation}
    S(x,y) = \sum_{t=1}^{n_q} B_t(x,y)\,,
\end{equation}
Assuming no motion, all photons incident at $(x,y)$ coming from the same scene point, which means $B_t(x,y)$ are i.i.d Bernoulli variables. Therefore $S$ follows a binomial distribution. The likelihood function for the unknown parameter $\phi$ given an observed number of photons $S=s$:
\begin{equation} \label{eq:likelihood}
    f(\phi\mid s) = \begin{pmatrix} n_q \\ n_q - s \end{pmatrix} (e^{-(\phi\tau\eta+r_q\tau)})^s (1-e^{-(\phi\tau\eta+r_q\tau)})^{n_q-s}\,.
\end{equation}
The maximum likelihood estimation (MLE) is given by:
\begin{equation} \label{eq:mle}
    \hat{\phi}= -\ln (1-s/n_q)/\tau\eta-r_q/\eta\,.
\end{equation}
The Fisher information can be computed as:
\begin{align}
    I(\phi) &= -\sum_{s=0}^N \frac{\partial^2}{\partial \phi^2}\log f(\phi\mid s)P\{S=s\}\\
    &= \frac{n_q \tau_q^2 \eta_q^2}{e^{\phi\tau_q\eta_q+r_q\tau_q}-1}\,.
\end{align}
For a sufficiently long sequence $n_q > 30$, the variance of the MLE can be estimated using Fisher information. Therefore,
\begin{equation}
    \rmse(\hat{\phi}_q) = \sqrt{\Var[\hat{\phi}_q]} \approx \frac{1}{\sqrt{I({\phi})}} = \sqrt{\frac{e^{\phi\tau_q\eta_q+r_q\tau_q}-1}{n_q \tau_q^2 \eta_q^2}}\,,
\end{equation}
which is consistent with the result in \cite{ingle_high_2019}.

\subsection{Auto-Exposure Strategy for Signal-to-Noise Ratio (SNR) Analysis}
Here we give the details of the strategy we use to determine the total exposure time and number of frames for the SNR analysis in Sec.~\mainSecAnalysis of the main paper. We only consider a single pixel for this analysis.
For a fair comparison, the total exposure time (sum of exposure time for all frames in the sequence) for both systems are assumed to be same, which is determined by
\begin{equation}
    T = \min(c_t/\phi, m_{max}/v)\,,
\end{equation}
where $c_t$ is a predetermined target count of photons. $\phi$ is the photon flux. $m_{max}$ is the maximum tolerable total motion in pixels. $v$ is the apparent speed of the pixel in pixels/s (we assume the speed is constant during the exposure). This strategy can be interpreted as attempting to choose an exposure time which allows us to record a target number $c_t$ of photons in the burst, while making sure that the total motion over the exposure doesn't exceed the set threshold $m_{max}$. In case the motion is too fast, the exposure time (and the number of photons recorded) is reduced proportionally so as to restrict it to $m_{max}$. This is because for too large apparent motion, the perfect alignment assumption usually does not hold -- due to brightness change, viewpoint change, or moving beyond the field of view. In practice, even if we take a long burst of images in this case, later images will not contribute to the merging due to matching difficulties.

In general, the number of frames $n_c$ for conventional burst photography is determined by balancing the motion blur and SNR of the resulting image: Choosing a larger $n_c$ will mitigate motion blur but also decrease SNR due to read noise. It is hard to compare the SNR of the methods when a single (conventional) frame contains motion blur. Therefore, we choose $n_c$ to be the minimum number that keeps the motion blur for a single frame below a certain threshold $m_f$ (\eg 1 pixel) and ignore the effects of motion blur when computing SNR.
\begin{equation}
    n_c = \frac{vT}{m_f}
\end{equation}
where $m_f$ is the maximum tolerable motion per frame. This is also similar to the auto-exposure strategy used in \cite{liba_handheld_2019}. In the analysis, we choose $c_t = 1000, m_{max} = 60$ (assuming a 512x256 camera), $m_f=1$.

For quanta burst photography, we always choose the maximum reachable frame rate since increasing the number of frames will not reduce SNR:
\begin{equation}
    n_q = \frac{T}{\tau_q}\,,
\end{equation}
where $\tau_q$ is the minimum frame time that is determined by the hardware.

\subsection{Dynamic Range Analysis} \label{subsec:analysis_dr}
\begin{figure}
  \centering
  \includegraphics[width=0.5\linewidth]{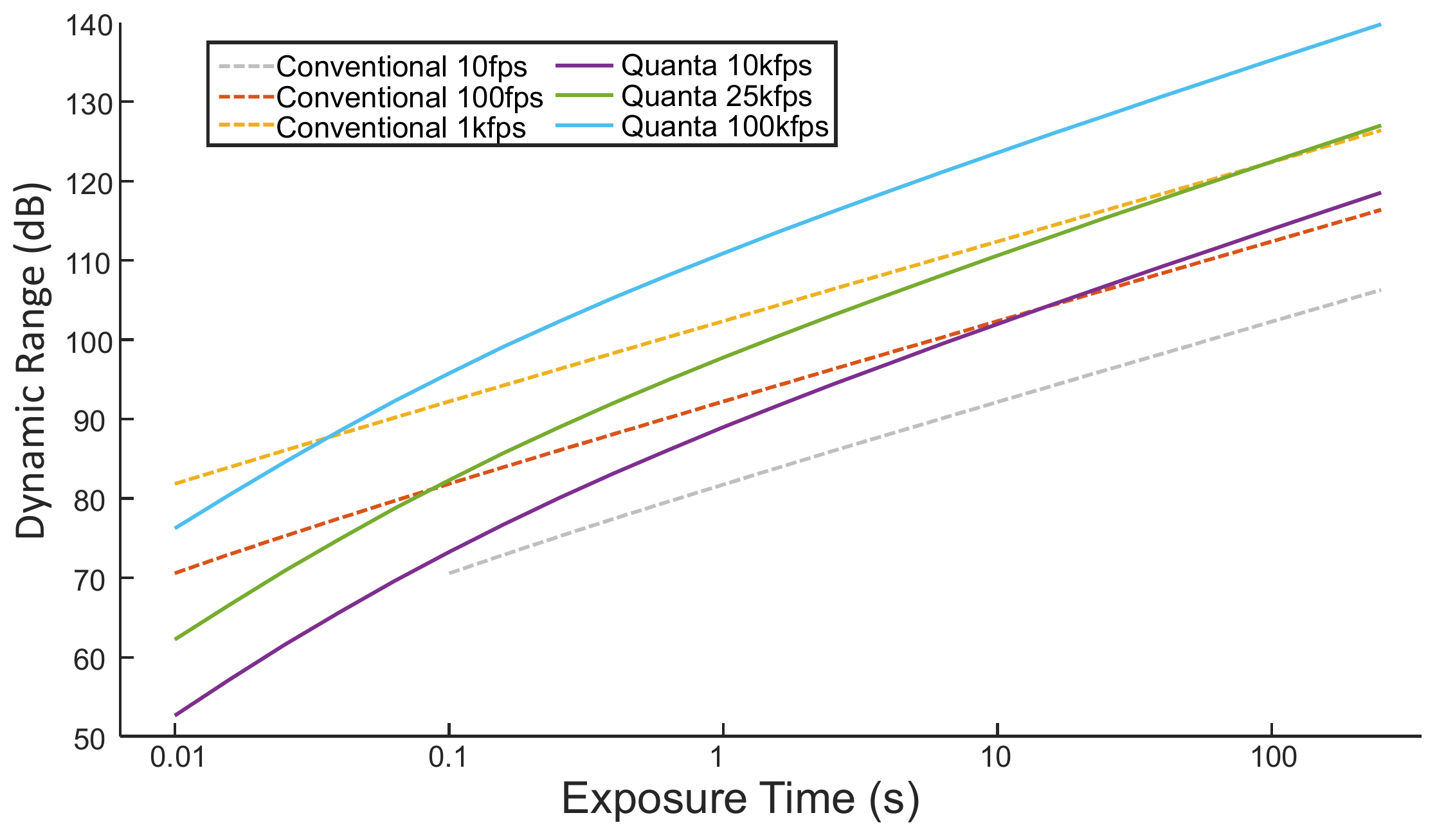}
  \caption{\textbf{DR analysis.} We plot the theoretical dynamic range of conventional burst photography and quanta burst photography as a function of exposure time. For both methods, we choose three typical frame rates. The dynamic range of quanta burst photography is lower due to low number of frames but grows very fast as the exposure time increases.}
  \label{fig:dr_analysis}
\end{figure}
In this section we give a theoretical analysis of the dynamic range of both conventional and quanta burst photography. We define the dynamic range as the ratio between the maximum measurable photon flux and the minimum measurable photon flux:
\begin{equation}
    DR = 20\log_{10}\frac{\phi_{max}}{\phi_{min}}
\end{equation}
where $\phi_{max}$ is defined as the highest photon flux before saturation. For conventional sensors, this is the case when the expected number of detected photons for each frame is equal to  $FullWellCapacity - 1$. For SPADs, this corresponds to the detection of $n_q-1$ photons in a total of $n_q$ frames,\ie $S=n_q-1$ in Eq.~\ref{eq:mle} .$\phi_{min}$ is defined as the lowest photon flux for which the SNR is above certain threshold. Here we choose the threshold to be 1 (0dB) which is consistent with previous works on SPAD~\cite{zarghami_high_2019,zappa_principles_2007} and common definition for conventional sensors.

Fig.~\ref{fig:dr_analysis} shows the dynamic range for conventional and quanta burst photography for different exposure time. The curves are plotted for a few typical frame rates for both image sensors. Quanta burst photography tends to perform worse for a short exposure time due to the low number of frames (low full well capacity), but grows faster than conventional burst photography as the exposure time increases. For example, 100kfps quanta burst photography performs better than 1kfps conventional burst photography as long as the exposure time is longer than 0.04s. \smallskip

\begin{figure}[t]
  \centering
  \includegraphics[width=0.9\linewidth]{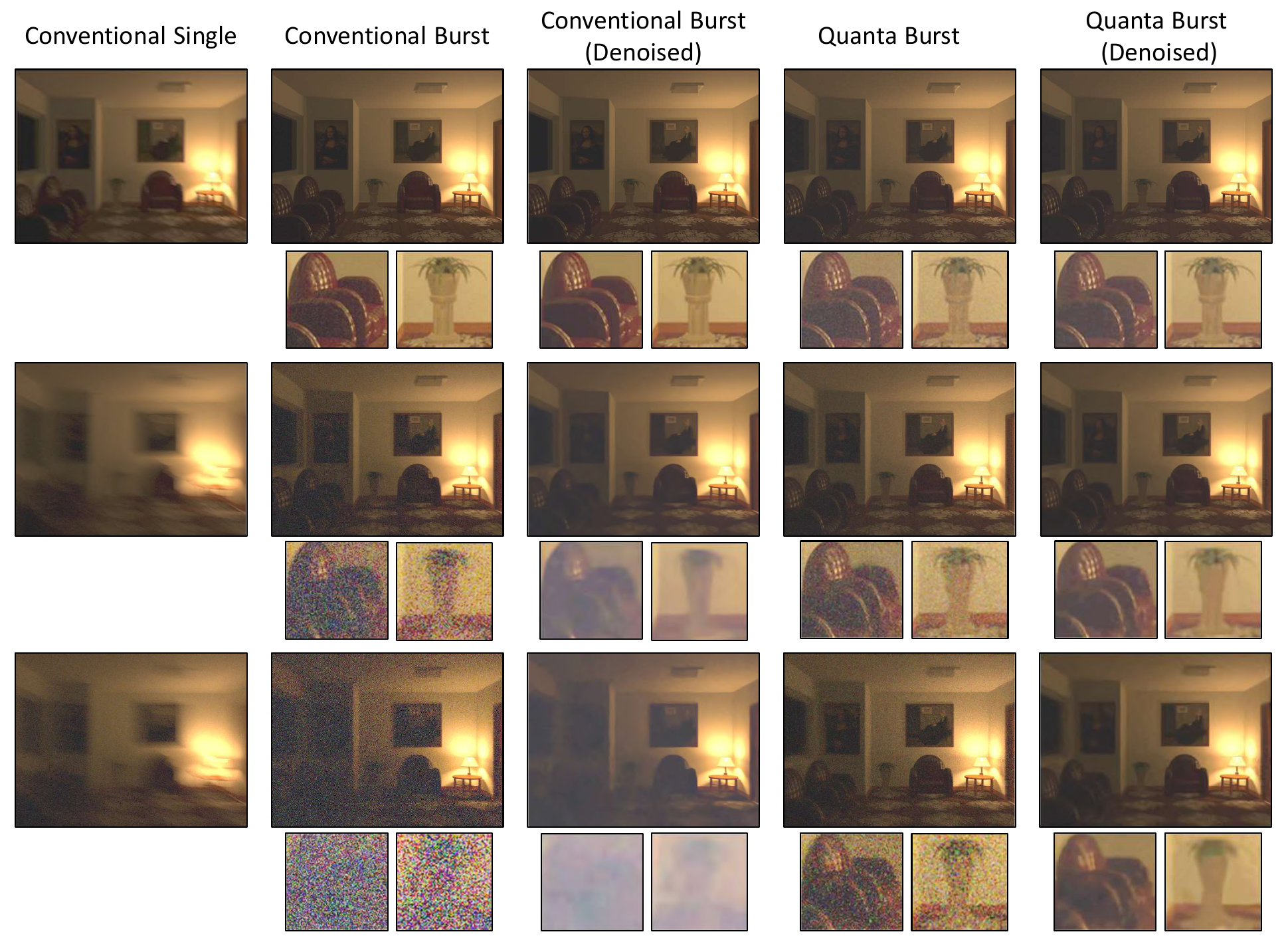}
  \caption{\textbf{Simulation results for different camera moving speeds.} We simulate the indoor scene with three different camera moving speeds. The camera moves linearly (with perturbations) at different speeds. \textbf{(Top)} When the camera motion is slow, the exposure time is chosen to meet a target number of total  collected photons (1000), in which case both methods generate high-quality images. The quanta burst photography gives slightly worse results due to higher dark current noise. \textbf{(Middle)} As the camera speed increases, the total exposure time is limited by the maximum tolerable apparent motion and a smaller number of photons are collected. As a result, the performance of both methods deteriorates. \textbf{(Bottom)} In the extremely fast case, the quanta burst photography can still recover the overall structure of the objects, while the conventional burst photography is completely dominated by noise.}
  \label{fig:sim_motion_speed}
\end{figure}

\begin{figure}[t]
  \includegraphics[width=\linewidth]{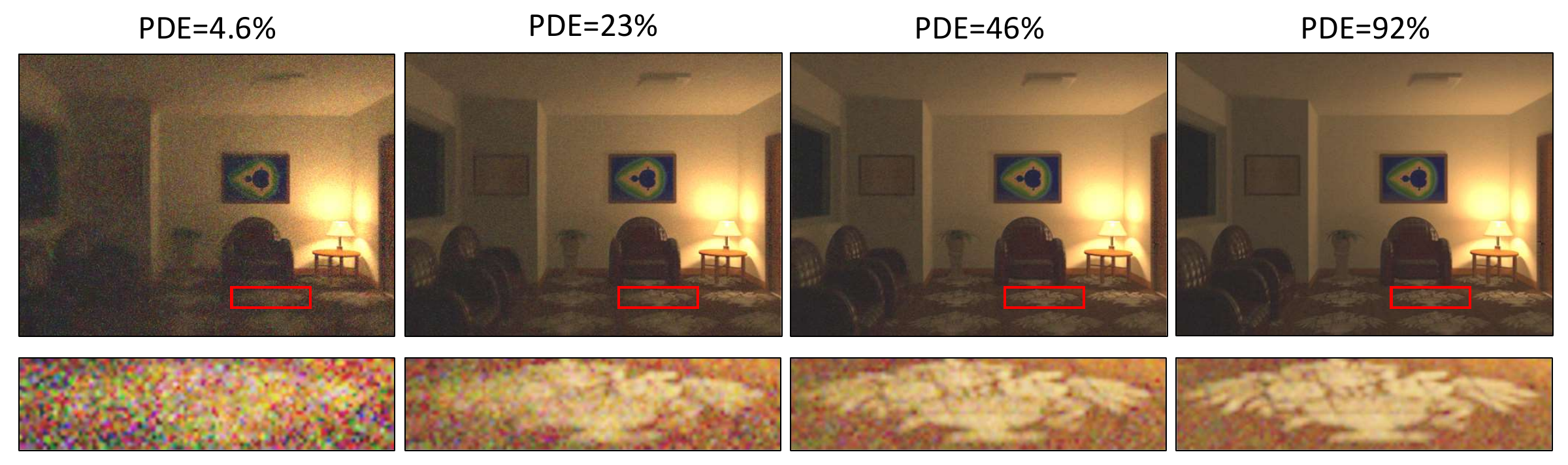}
  \caption{\textbf{Performance for different PDEs.} We show the reconstruction results for the same scene under same camera motion for different PDE. The figure titles show the PDE of the green channel, which correspond to current specification without microlens, current specification with microlens, double fill factor, double fill factor and double PDP. PDE is an essential factor of the final image quality, as shown in the close-ups.}
  \label{fig:sim_qe}
\end{figure}

\begin{figure}[t]
  \includegraphics[width=\linewidth]{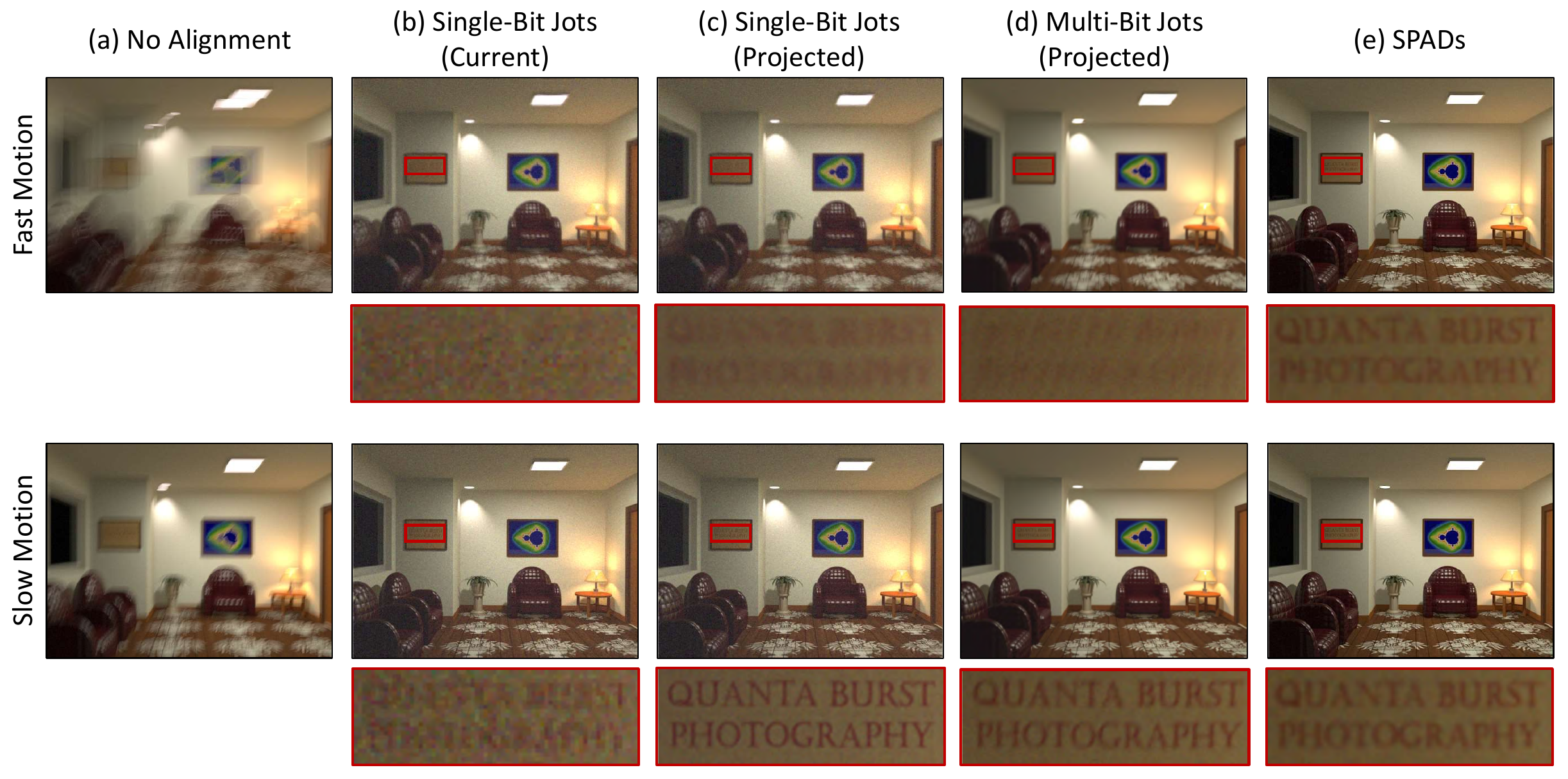}
\caption{\textbf{Comparison of jots and SPADs under different motion speeds.} Current implementation of jots perform worse than SPADs for both fast and slow motion. We simulate a projected jots which is assumed to have the same bandwidth as SPADs and work in both single-bit and 4-bit mode. For fast motion, temporally-supersampled SPADs are able to resolve the motion blur and give sharper image. For slow motion, spatially-supersampled jots are able to reconstruct better image details. Multi-bit jots generate slightly blurred image due to lower frame rate. We expect SPADs and jots to complement each other and work for different motion ranges.}
\label{fig:sim_jots_motion_more}
\end{figure}

\begin{figure}[t]
  \includegraphics[width=\linewidth]{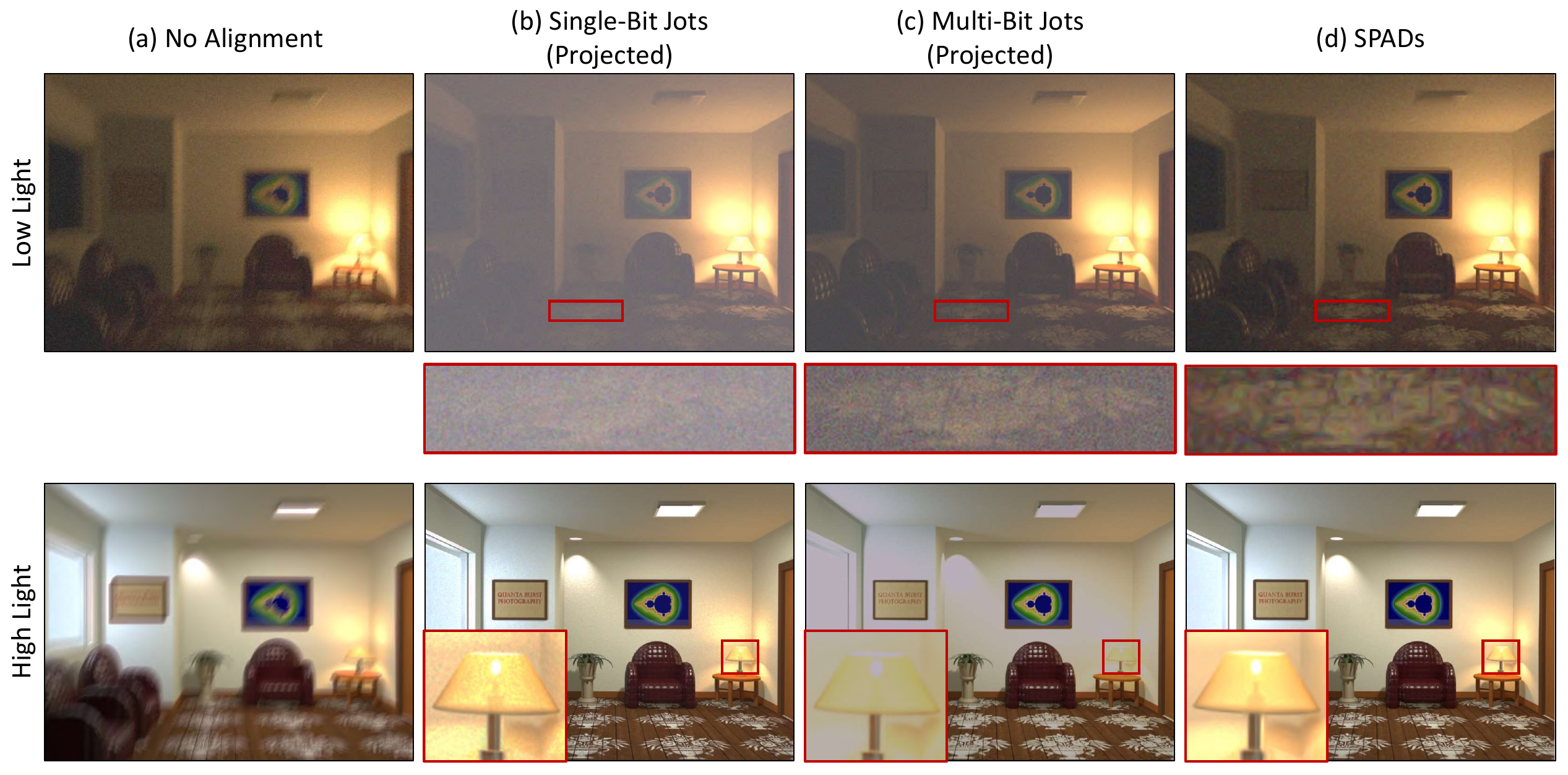}
\caption{\textbf{Comparison of jots and SPADs under different lighting conditions.} In a dark environment, single-bit jots contain significant noise. Multi-bit jots contain less noise due to its longer exposure time and more discrete light levels. SPADs contain least noise since there is no read noise. In a bright scene, single-bit jots are near to saturation. Multi-bit jots are saturated and the image appears washed out. SPADs are able to reconstruct the high flux scene points.}
\label{fig:sim_jots_lighting_more}
\end{figure}



\begin{figure}[t]
  \includegraphics[width=\linewidth]{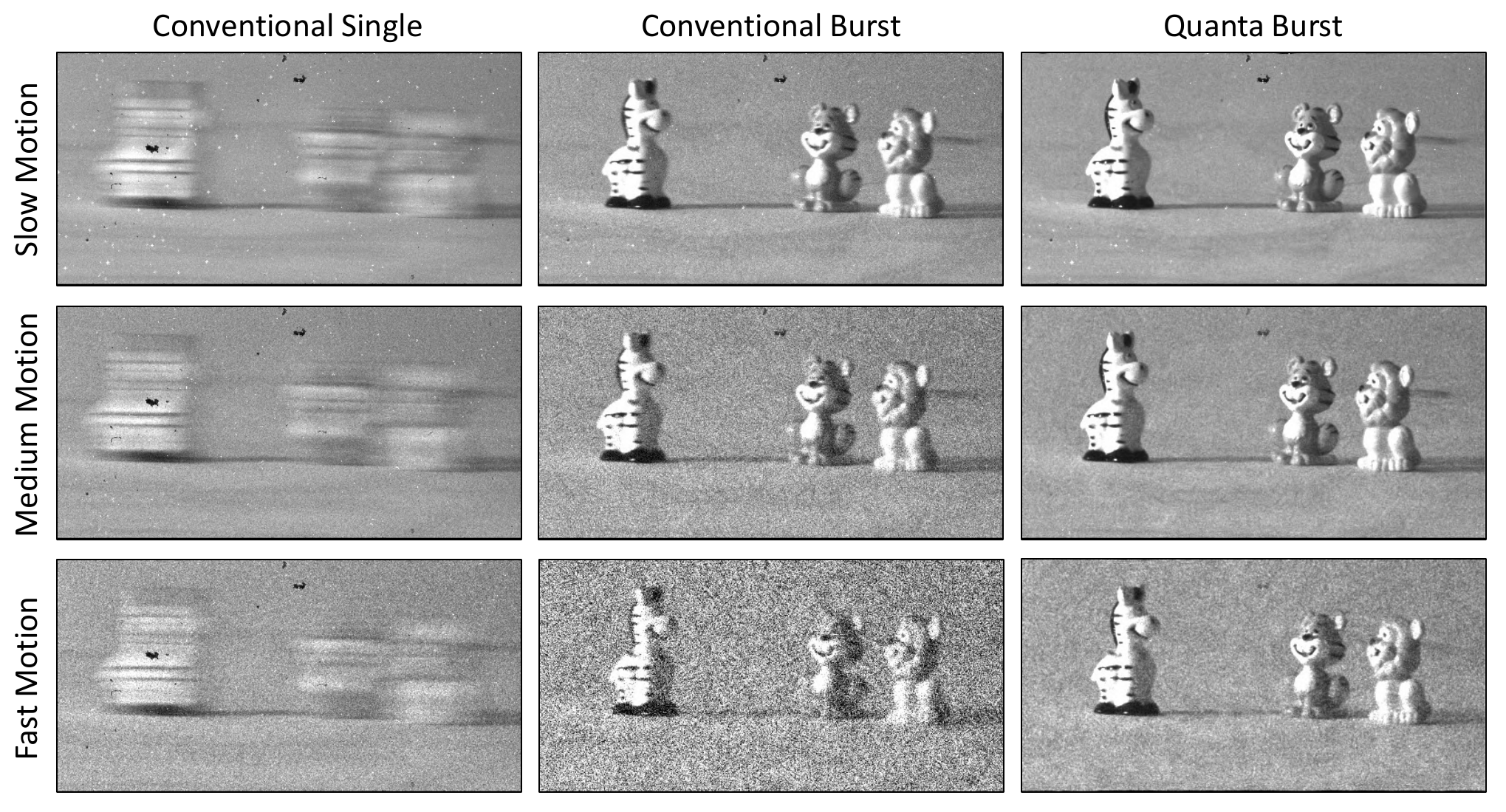}
  \caption{\textbf{Performance for different camera moving speeds.} We capture three binary sequences for the same scene with the camera moving at different speeds. For fast motion, conventional cameras can generate either a heavily blurred image or an image with significant noise, while quanta burst photography can reconstruct a blur-free image with much lower noise.}
  \label{fig:real_motion_speed}
\end{figure}

\begin{figure}[t]
  \centering
  \includegraphics[width=0.97\linewidth]{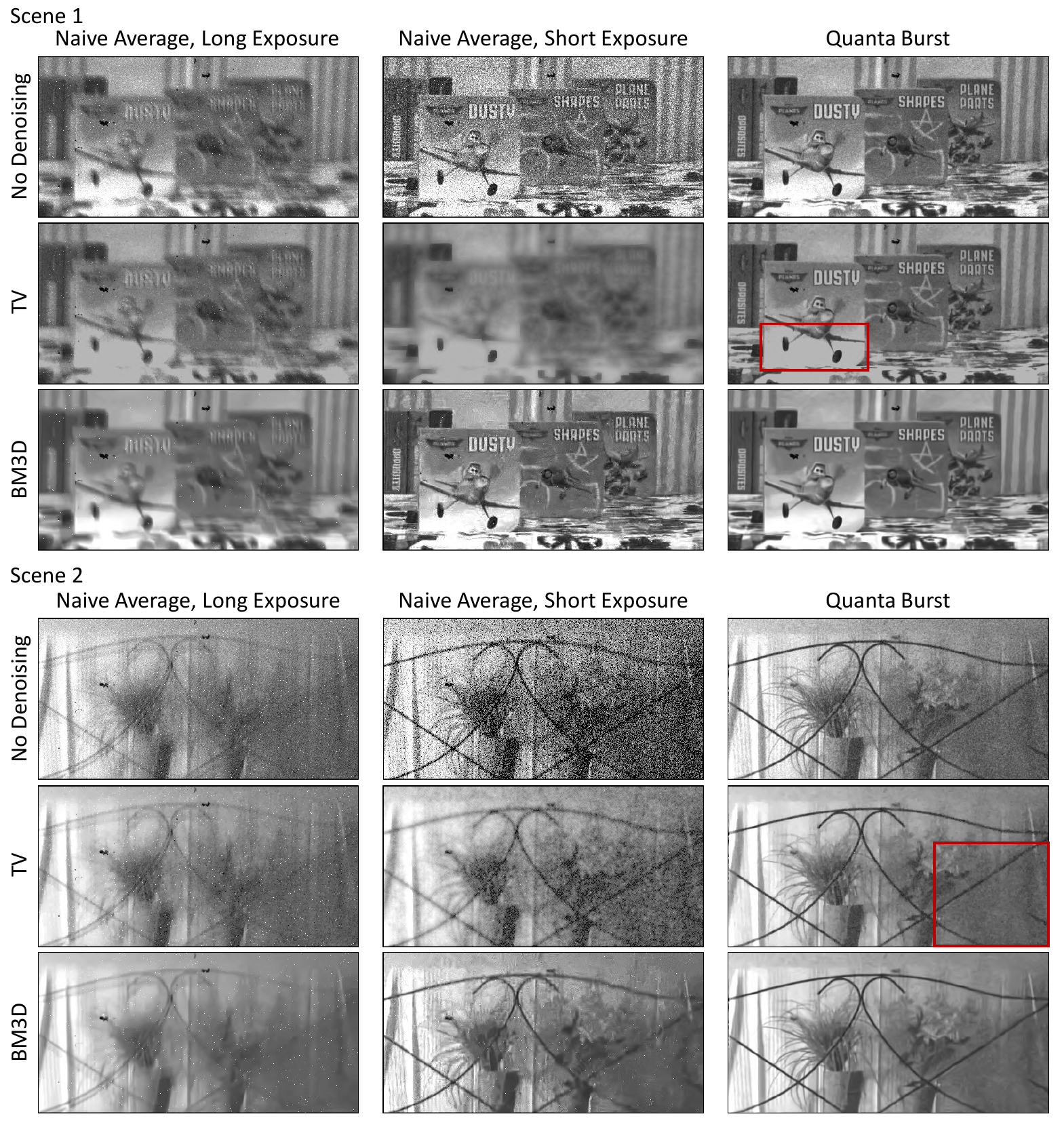}
  \caption{\textbf{Comparison of denoising algorithms.} \textbf{(Left)} Naive average reconstruction without motion compensation on a long sequence (200 images). Results contain severe motion blur. \textbf{(Center)} Naive average reconstruction without motion compensation on a short sequence (20 images). Results are sharp but contain a lot of noise. Applying denoising algorithms help reduce noise but also remove high-frequency image details. \textbf{(Right)} Burst align and merge results on 200 images. Results are sharp and less noisy. Applying denoising algorithms further reduce noise. BM3D performs better than TV as TV reduces intensity contrast in bright regions (Scene 1) and does not reduce noise well in dark regions (Scene 2).}
  \label{fig:exp_denoise}
\end{figure}

\begin{figure}[t]
  \centering
  \includegraphics[width=0.97\linewidth]{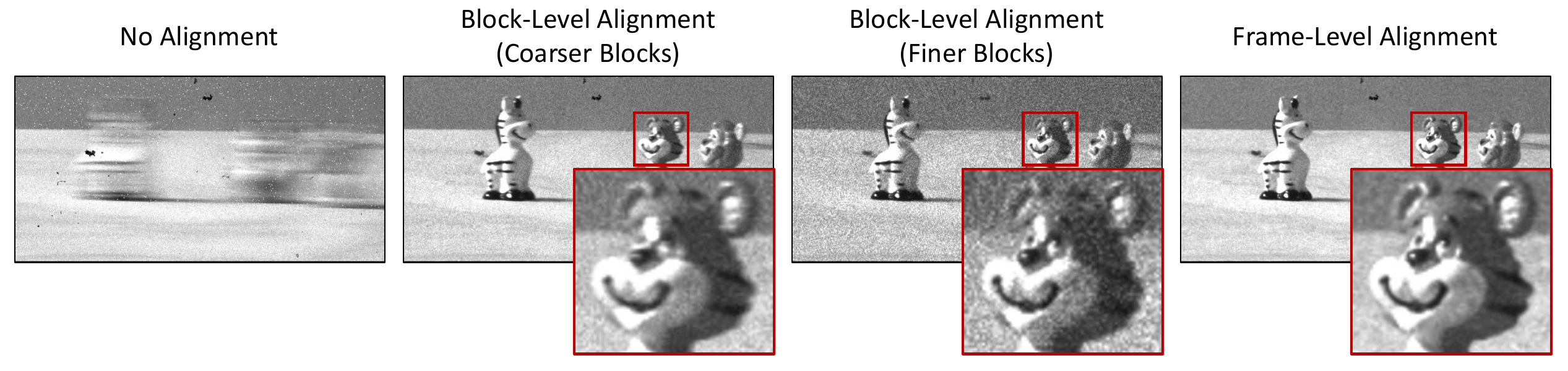}
  \caption{\textbf{Effects of frame-level flow interpolation.} A short sequence that contains 2000 binary frames, which is divided into 100-frame blocks (coarser blocks) and 20-frame blocks (finer blocks). Results from blocker-level alignment either contain motion blur (coarser blocks) or more noise (finer blocks), while the interpolated frame-level alignment is able to remove motion blur without increasing the amount of noise.}
  \label{fig:exp_interp}
\end{figure}

\begin{figure}[t]
  \includegraphics[width=\linewidth]{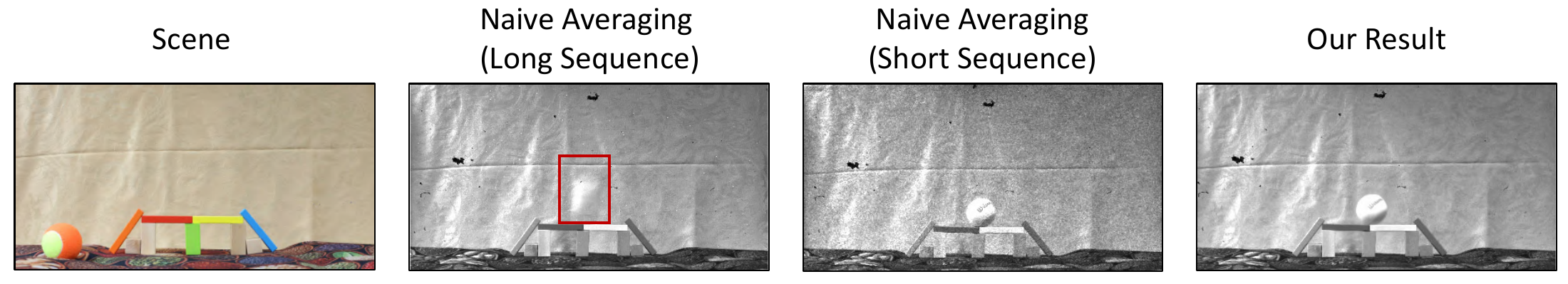}
  \caption{\textbf{Resolving scene motion.} We capture a binary image sequence where a tennis ball is dropped vertically. Quanta burst photography is able to align the images to generate a blur-free image with high SNR.}
  \label{fig:real_tennis}
\end{figure}

\begin{figure}[t]
  \includegraphics[width=\linewidth]{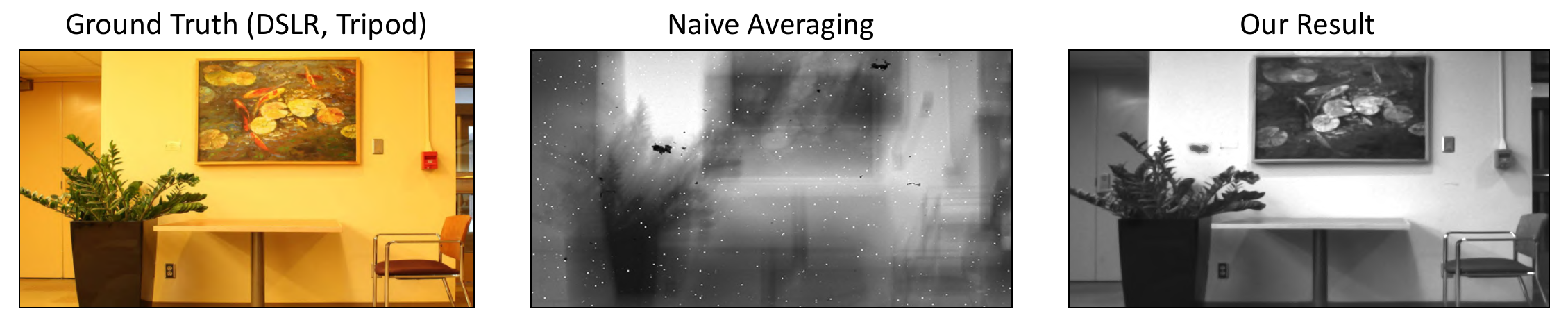}
  \caption{\textbf{Indoor scene under natural lighting.} The binary sequence is correctly aligned to each other despite intense camera motion, resulting in a clear, sharp image.}
  \label{fig:real_indoor_more}
\end{figure}

\section{Results}
\subsection{Simulation Results}
The simulated images in the main paper and supplementary material is rendered using POV-Ray. Code is adapted from Jaime Vives Piqueres' Lightsys demo\footnote{http://www.ignorancia.org/index.php/technical/lightsys/}. 

\smallskip\noindent\textbf{Comparison of conventional burst photography and quanta burst photography.} 
Fig.~Supp-\ref{fig:sim_motion_speed} shows the simulation results for three different motion speeds. According to the strategy in Sec.~\ref{sec:analysis}, when the apparent speed gets faster, the total exposure time is limited by the maximum tolerable amount of motion to avoid appearance changes due to significant viewpoint change, which results in a smaller number of incoming photons. Similar to Fig.~\mainFigSimLighting in the main paper, the quality of conventional burst result goes down faster than quanta burst result.
The results of these two sets of simulation match the theoretical analysis of SNR in Sec.~\mainSecAnalysis of the main paper: Quanta burst photography performs better in low light and fast moving scenarios.

\smallskip\noindent\textbf{Performance for different photon detection efficiency (PDE).} Fig.~Supp-\ref{fig:sim_qe} shows the reconstruction results with different assumed PDE of the single-photon camera, which corresponds the current specification without microlens, with microlens, double fill factor than current specification, and double fill factor and double PDP than current specification. PDE is an essential factor that determines the final image quality. We expect PDE of SPAD cameras to keep improving due to improving fabrication processes.

\smallskip\noindent\textbf{Comparison of jots and SPADs.} Fig.~Supp-\ref{fig:sim_jots_motion_more} shows the comparison between jots and SPADs-based quanta burst photography. The proposed quanta burst photography is adapted to single-bit and multi-bit jots. The input to the align process is not binary images but spatially downsampled versions of single-bit and multi-bit images (using a box filter, normalized in 0-1). The rest of the pipeline still works with this data format. After merging, a linear response function is applied to recover the intensities for multi-bit jots~\cite{gnanasambandam_megapixel_2019}.

As mentioned in the main paper, because of the limited spatial and temporal resolution, current jots perform worse than SPADs. The projected single-bit and multi-bit jots are not able to remove the motion blur for extremely fast motion. For slow motion, they are able to generate sharper images than SPADs thanks to higher spatial resolution. Multi-bit jots generate slightly blurred images due to their lower frame rate.

Fig.~Supp-\ref{fig:sim_jots_lighting_more} shows the comparison between jots and SPADs in extremely low and high lighting conditions. In the low light condition, single-bit jots images are noisy because the read noise, albeit deep sub-electron, makes the pixels flip between 0 and 1. In such low lighting, most pixel receive 0 photon during the exposure. Therefore, more pixels are flipped from 0 to 1, resulting in a whitened, noisy image. The result cannot be improved by setting a threshold larger than one photon, since few pixels receive more than 1 photon. Multi-bit jots generate better result because the signal is stronger compared to read noise due to the longer exposure and higher full well capacity. SPADs contain least noise since there is no read noise.

In the high light condition, 4-bit jots saturate more easily than 1-bit jots. This is because single-bit jots have a nonlinear response curve. Increasing the number of bit will decrease the overexposure latitude and result in a more linear response curve~\cite{fossum_modeling_2013}. It is not clear whether such nonlinearity can be used to extend the dynamic range for multi-bit reconstruction, which is also likely to be limited by the non-uniformity of the jots. Here we follow the practice in \cite{gnanasambandam_megapixel_2019} and use a linear response function for reconstructing multi-bit images. 



Notice that all the analysis above assumes bandwidth is the bottleneck for all types of sensors. In practice, frame rate and spatial resolution may be constrained by other factors in chip design and manufacture. The analysis is a based on current specification of jots and SPADs. In the future, jots may be 
able to achieve read noise lower than 0.15e$^{-}$ which will result in improved dynamic range.



\subsection{Experimental Results}
\noindent\textbf{Performance for different camera moving speeds.}
Fig.~Supp-\ref{fig:real_motion_speed} shows the performance of the proposed method for different camera moving speeds. Same as Fig.~\mainFigRealLighting in the main paper, the conventional images are simulated by reconstructing intensities from binary frames and then adding read noise and quantization error. As the camera moves faster, the sensors collect a lower number of photons, and the results for both methods degenerate. In the fastest scenario, conventional camera captures images with either significant blur or low SNR, while quanta burst photography is able to resolve the motion and achieve an acceptable SNR.

\smallskip\noindent\textbf{Comparison of single-photon imaging denoising algorithms.} In this paper we focus on combining information from all other auxiliary frames in a sequence to help denoise the reference frame. After merging all frames into a single sum image, it is still possible to apply single-image denoising and reconstruction algorithm, using spatial information to further improve the SNR of the image. Fig.~Supp-\ref{fig:exp_denoise} (right) show the results of applying two denoising algorithms after burst merging: BM3D and total variation (TV). BM3D is applied as a post-processing step after Anscombe transform, as noted in the main paper. Total variation is formulated as a joint reconstruction-denoising optimization problem~\cite{chan_efficient_2014}:
\begin{equation}
    \min_{\vec{\phi}} -\sum_{i\in\Omega}\log f(\phi_i \mid s_i) + \lambda_{tv} \Vert \mat{D}\vec{\phi} \Vert_1 \,,
\end{equation}
where $\vec{\phi}$ is a vector representation of the photon flux at each pixel. $\Omega$ is the image domain. $f(\phi_i \mid s_i)$ is the likelihood function defined in Eq.~\ref{eq:likelihood}. $\mat{D}$ is the finite difference operator that is used to compute the gradients. $\lambda_{tv}$ is a parameter used to control the amount of spatial smoothing.

The two sequences are temporally subsampled from the original sequences in Fig.~\mainFigRealChallenge in the main paper, which contain only 200 binary images and therefore the results are much noisier. We notice that in general BM3D performs better than TV. In Fig.~Supp-\ref{fig:exp_denoise} (Scene 1, Burst), TV is not able to preserve the contrast in the region indicated by the red rectangle. In (Scene 2, Burst), the darker region in the scene is noisier in TV than in BM3D.

We also compare the result of the proposed quanta burst photography with directly denoising a simple average of the binary sequence as is done in previous papers (without compensating for motion). Since the scene is moving, the naive average results are either with heavy motion blur (long sequence), or contain a lot of noise (short sequence). By comparing the naive result with short sequence and burst result, it is clear that using temporal information for denoising helps remove noise while keep the spatial details of the image.

\smallskip\noindent\textbf{Effects of frame-level flow interpolation.} One of the main technical contributions of the proposed quanta burst photography method is that the alignment is performed on aggregated block sum images and is then interpolated to obtain frame-level patch flow, which is later used for merging. Fig.~Supp-\ref{fig:exp_interp} shows how the frame-level patch flow interpolation helps resolve motion blur. 
If we divide the 2000-frame sequence into 100-frame blocks (``coarser blocks'', 20 blocks in total) and do not interpolate the flow within the block (\ie merge the block sum image directly) , the result contains noticeable motion blur. The motion blur can be resolved by dividing into finer blocks (20 frames per block $\times$ 100 blocks), but this results in a noisier image. This is because more noise is preserved when a smaller block size is used, as discussed in Sec.~\mainSecMerge in the main paper. Frame-level flow interpolation (interpolated from coarse block division) is able to remove the motion blur while not adding extra noise.


\smallskip\noindent\textbf{Resolving scene motion.} Fig.~Supp-\ref{fig:real_tennis} shows another sequence with scene motion: A tennis ball is dropped vertically. Naive averaging of the binary sequences results in either motion blur or lots of shot noise. By aligning the image patches that constitute the tennis ball properly, the proposed method is able to generate a high-SNR image without motion blur.

\smallskip\noindent\textbf{Indoor scene with natural lighting.} Fig.~Supp-\ref{fig:real_indoor_more} shows another sequence of indoor scene with natural lighting. The intense camera motion between the frames is correctly resolved and a clear, sharp image is generated.

\bibliographystyle{ieeetr}
\bibliography{supp}